\documentclass[runningheads]{llncs}

\usepackage{accv}

\usepackage{accvabbrv}

\usepackage{graphicx}
\usepackage{booktabs}
\usepackage{makecell}
\usepackage{caption}
\usepackage{gensymb}
\usepackage{array}

\usepackage[accsupp]{axessibility}  

\usepackage[pagebackref,breaklinks,colorlinks,citecolor=accvblue]{hyperref}

\usepackage{orcidlink}

\begin{document}

\title{GameIR: A Large-Scale Synthesized Ground-Truth Dataset for Image Restoration over Gaming Content} 

\titlerunning{GameIR}

\author{
    Lebin Zhou\inst{1,2} \and
    Kun Han\inst{2} \and
    Nam Ling\inst{1} \and
    Wei Wang\inst{2} \and
    Wei Jiang\inst{2}
}


\authorrunning{L.~Zhou et al.}

\institute{
    Santa Clara University \email{\{lzhou,nling\}@scu.edu} \and
    Futurewei Technologies Inc. \email{\{lzhou,khan,rickweiwang,wjiang\}@futurewei.com}
}

\maketitle

\begin{abstract}
Image restoration methods like super-resolution and image synthesis have been successfully used in commercial cloud gaming products like NVIDIA’s Deep Learning Super Sampling (DLSS). However, restoration over gaming content is not well studied by the general public. The discrepancy is mainly caused by the lack of ground-truth gaming training data that match the test cases. Due to the unique characteristics of gaming content, \textit{e.g.}, sharp and clear low-resolution (LR) images, the common approach of generating pseudo training data by degrading the original high-resolution (HR) images results in inferior restoration performance. 
In this work, we develop GameIR, a large-scale high-quality computer-synthesized ground-truth dataset to fill in the blanks, targeting at two different applications. The first is super-resolution with deferred rendering, to support the gaming solution of rendering and transferring LR images only and restoring HR images on the client side. We provide 19200 LR-HR paired ground-truth frames coming from 640 videos rendered at 720p and 1440p for this task. The second is novel view synthesis (NVS), to support the multiview gaming solution of rendering and transferring part of the multiview frames and generating the remaining frames on the client side. This task has 57,600 HR frames from 960 videos of 160 scenes with 6 camera views (with associated camera intrinsic and extrinsic parameters). In addition to the RGB frames, the GBuffers during the deferred rendering stage (\textit{i.e.}, segmentation maps, and depth maps) are also provided, which can be used to help restoration. Furthermore, we evaluate several SOTA super-resolution algorithms and NeRF-based NVS algorithms over our dataset, which demonstrates the effectiveness of our ground-truth GameIR data in improving restoration performance for gaming content. Also, we test the method of incorporating the GBuffers as additional input information for helping super-resolution and NVS. We release our dataset and models to the general public to facilitate research on restoration methods over gaming content.

\end{abstract}

\section{Introduction}

Modern cloud gaming has become increasingly popular with an expected global market share value reaching over \$12 billion by 2025. By streaming frames rendered on remote servers to users' devices, cloud gaming benefits both users and game developers. Users can play a large library of games on any device, without requiring expensive hardware, and game developers can optimize games for known server-side hardware, without dealing with heterogeneity of client-side devices. In recent years, Generative AI (GAI) technologies, such as GAN and diffusion models, are transforming the gaming industry by enabling fast and accessible high-quality content creation. GAI makes it possible for anyone to build and design games without professional artistic and technical knowledge, further empowering immeasurable market growth.

Being the next-generation game changer, cloud gaming poses tremendous challenges for data compression and transmission. Most current solutions rely on heavy server-side computation and network delivery, where the client device is merely used for display. It is difficult for a client to enjoy high-quality gaming if the bandwidth is limited, even with a powerful client device. To avoid input delay and over-consuming bandwidth, high-quality frames need to be heavily compressed with extremely low latency. Traditional codecs like H.264/H.265/H.266 \cite{vvc_std,hevc_std} or recent neural video coding \cite{DVC2019} targeting natural videos cannot resolve this transmission bottleneck.

Generative methods like GAN, when applied to super-resolution and image rendering and synthesis, can largely alleviate the transmission issues. Server-side computation and transmission can be reduced by leveraging the computation power of client devices. For example, the server can render low-resolution (LR) frames to transfer, and high-resolution (HR) frames can be computed on the client side. In multiview (\textit{e.g.}, immersive VR) gaming, the server can render part of the frames or views to transfer, and the remaining frames or views can be computed by client devices. NVIDIA's Deep Learning Super Sampling (DLSS) technology \cite{NVDLSS1, NVDLSS2, NVDLSS3} has commercialized this idea, demonstrating the great potential of optimizing the gaming experience by leveraging bandwidth conditions and computation power of client devices.

The key factor of the success of DLSS is the large-scale ground-truth LR-HR paired data or multiview gaming data used for training that matches the test scenarios. In comparison, the research community uses pseudo training data for many restoration tasks \cite{IRreview1, IRreview2, IRreview3, VRreview1}. For example, for super-resolution, the LR data is generated from the HR data by downsampling and adding degradation-like noises, blurs, and compression artifacts. Such pseudo training data does not match the real gaming data. For example, as shown in Fig.~\ref{fig:gaming_characteristics}, true LR gaming frames are high-quality, sharp, and clear without noises or blurs, different from generated pseudo LR data. Also, there are unnatural visual effects and object movements, but with little motion blur, different from captured natural videos. As a result, we have to resort to ground-truth gaming data for effective training. Unfortunately, it is non-trivial to obtain such ground truth, which requires technical skills, labor, and computation using graphics engines. 

In this paper, we provide GameIR, a large-scale computer-synthesized ground-truth dataset to facilitate the research of restoration methods over the gaming content. We aim to bring the success of commercial-level DLSS to the public research community so that AI-empowered cloud gaming solutions using image restoration techniques can be more effectively investigated in the field. Our contributions can be summarized as follows.

\begin{itemize}
    \item 
    We develop a large-scale, high-quality, computer-synthesized ground-truth dataset aiming at two different applications: super-resolution with deferred rendering to support the gaming solution of transferring LR images and restoring HR images on the client side, and novel view synthesis (NVS) to support the gaming solution of transferring part of the multiview frames and generating the remaining frames on the client side. For super-resolution, the GameIR-SR dataset contains 19,200 LR-HR paired ground-truth frames derived from 640 videos rendered at 720p and 1440p using the open-source CARLA simulator with the Unreal Engine. In addition to the LR-HR paired RGB images, the additional GBuffers during the deferred rendering stage (\textit{i.e.}, segmentation maps, and depth maps) are also provided. An example is shown in Fig.~\ref{fig:sr_dataset_example}. For NVS, the GameIR-NVS dataset contains 57,600 HR frames at 1440p from 960 videos of 160 scenes with 6 camera views (with associated camera intrinsic and extrinsic parameters). Besides multiview RGB frames, the dataset also includes the corresponding segmentation maps and depth maps. An example is illustrated in Fig.~\ref{fig:multiview_dataset_example}.

    \item 
    We evaluate over our dataset several existing SOTA algorithms. For super-resolution, we test Anime4K \cite{Anime4K} that is designed for anime/cartoon content, RealESRGAN \cite{Realesrgan} that learns real-world degradations for general images, and AdaCode \cite{AdaCode} that uses learned generative codebook priors. For NVS, we test Instant-NGP \cite{Instant-NGP}, NeRFacto \cite{nerfstudio}, DSNeRF\cite{DSNeRF} and PyNeRF\cite{pynerf}, which represent the latest NeRF-based NVS methods. We aim to provide a baseline to understand how current methods perform over real gaming data so that improved solutions can be further studied.

    \item 
    We further evaluate how models can benefit from the additional GBuffers. For super-resolution, GBuffers are either used as additional inputs by concatenating with the RGB frames or used as generative conditions during the conditional restoration process by feature modulation. For NVS, the depth map is used to assist NeRF-based models by providing additional geometry information for the scene.

\end{itemize}

We release our GameIR dataset and the evaluated models. To the best of our knowledge, this is the first large-scale video set providing ground-truth computer-synthesized LR-HR paired or multiview frames with associated GBuffer data at the scene level. Our dataset can help to advance the research of restoration methods over gaming content for the general public. 

\section{Related Works}

\subsection{Super-Resolution: Methods and Datasets}
The pioneering work of SRCNN \cite{SRCNN1, SRCNN2} has inspired extensive research on deep-learning-based single image super-resolution (SISR). Earlier works \cite{johnson2016perceptual, kim2016accurate, zhang2017learning, lim2017enhanced, zhang2018residual, wang2018esrgan} assumed that LR images were generated by ideal degradation models. To handle the complex real degradations consisting of unknown factors like blurring and noises, blind SISR \cite{michaeli2013nonparametric, zhang2018learning} has become the research focus, which can be categorized as explicit and implicit methods. Explicit methods \cite{zhang2018learning, gu2019blind, zhang2020deep, maeda2022image} explicitly model and estimate the degradation process, and perform reconstruction based on the estimated degradation model. However, real-world degradation is too complicated to model accurately through simple combinations of multiple degradations. In comparison, implicit methods \cite{ESRGAN, swinIR, Realesrgan, AdaCode, BSRGAN, Anime4K} automatically learn and adapt to various degradation conditions based on LR training data distribution. Although implicit methods have achieved large improvements over real-world images, their performance is highly limited by the training degradations, making them difficult to generalize to out-of-distribution images. Previous methods mitigate this issue by increasing the variety of training degradation types and scales. However, such a training strategy does not work well for gaming content. Real-rendered LR images are clear, sharp, and without blur or artifacts, which are quite different from pseudo LR images generated by applying degradations. As a result, SISR models need to be trained on real LR-HR paired gaming data to learn true degenerative features and improve their performance.

Existing datasets for SISR mainly consist of HR images. Commonly used datasets include DIV2K \cite{DIV2K}, DIV8K \cite{DIV8K}, Flickr2K \cite{Flickr2K}, and Flickr-Faces-HQ (FFHQ) \cite{FFHQ}. Other popular datasets for general vision tasks, such as ImageNet \cite{ImageNet} and COCO \cite{COCO}, are also used for SISR. LR images for these datasets are generated by applying degradations to the HR images. To improve the generalization of models when applied to real-world scenarios, complex degradations have been employed, such as multiple simulated degradations \cite{elad1997restoration} and BSRGAN generated degradations \cite{BSRGAN}. However, the gap between the simulated and real degradations still exists. The problem is especially prominent for gaming images due to their unique characteristics different from natural images. 

There are some datasets providing real-world ground-truth image pairs, \textit{e.g.}, City100 \cite{City100}, RealSR \cite{RealSR}, and DRealSR \cite{DRealSR}, by using two calibrated devices with varying focal lengths to directly capture LR-HR image pairs. However, due to the expensive process, scale and content diversity is usually highly limited. Also, time synchronization and pixel-level alignment still remain challenging. 

\subsection{Novel View Synthesis: Methods and Datasets}

NVS aims to generate novel view images by integrating image data from multiple camera perspectives. Recently, methods based on Neural Radiance Fields (NeRF) \cite{NeRF} have shown great performance over a large variety of scenes.  NeFR++ \cite{NeRF++} builds upon NeRF with improved representations and volume rendering. Mip-NeRF \cite{Mip-NeRF} uses conical frustums rendering to reduce aliasing and enhance applicability to multiscale and high-resolution scenes. Instant-NGP \cite{Instant-NGP} uses hash tables and multi-resolution grids to speed up training and inference. DSNeRF \cite{DSNeRF} leverages depth information for supervision to improve performance. 3D Gaussian Splatting (3DGS) \cite{Splatting} uses Gaussian functions for real-time, high-quality rendering. PyNeRF\cite{pynerf} enhances the rendering speed and quality by training models across various spatial grid resolutions.

NVS datasets are generally divided into synthetic and real-world.  Most synthetic datasets are at the object level. 
Blender \cite{NeRF}, Objaverse \cite{Objaverse}, and D-NeRF \cite{D-nerf}  are typical synthetic datasets, which contain 3D CAD models with varied textures and geometries without real-world noises or non-ideal conditions. 

Earlier real-world datasets were developed for multi-view stereo tasks, such as Tanks and Temples \cite{tanks-and-temples} and DTU \cite{DTU}, offering limited scene variety. ScanNet \cite{Scannet} contains 3D scans and RGB-D video data, but with motion blur and narrow field-of-view. Later datasets featuring outward-facing and forward-facing scenes have limited diversity in general. For instance, LLFF \cite{llff} provides 24 cellphone-captured forward-facing scenes. Mip-NeRF 360 \cite{Mip-NeRF} provides 9 indoor and outdoor
scenes with uniform distance around central subjects. Mill 19 \cite{Mill-19} provides 2 industrial and open-space scenes. 
BlendedMVS \cite{Blendedmvs} offers multi-view images and depth maps but with limited scenes. Recently, large-scale scene-level real-world datasets have emerged. For example, RealEstate10K \cite{RealEstate10K} offers diverse indoor scenes through real estate videos, but with low-resolution and inconsistent quality. Replica \cite{Replica} provides high-quality data including RGB images, depth maps, and semantic annotations, but is limited to indoor environments only. The most recent DL3DV-10K \cite{DL3DV-10K} significantly enriched the real-world scene collection by providing 10,510 videos captured from 65 types of scene locations, with different levels of reflection,
transparency, and lighting conditions. 

In comparison, there is a lack of large-scale scene-level synthetic datasets for NVS research over synthetic gaming data. Similar to the super-resolution task, due to the unique characteristics of gaming content, \textit{e.g.}, unnatural object motion with limited motion blur, NVS methods need to be trained and evaluated over scene-level synthetic datasets to assess their effectiveness for gaming data. 

\section{Our Dataset}

In this work, we develop the GameIR dataset, which is a large-scale synthetic scene-level dataset to facilitate image restoration research for cloud gaming solutions. GameIR provides ground-truth LR-HR pairs and synchronized multiview video frames to support both super-resolution and NVS tasks. 

\subsection{Acquisition Environment and Settings}

\begin{figure}[htbp]
\centering

\begin{subfigure}[b]{0.24\textwidth}
    \includegraphics[width=\textwidth]{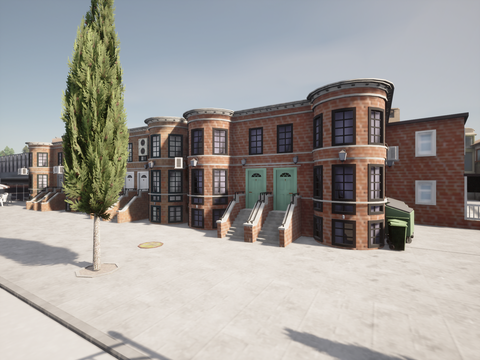}
    \caption{Town01}
\end{subfigure}
\hfill 
\begin{subfigure}[b]{0.24\textwidth}
    \includegraphics[width=\textwidth]{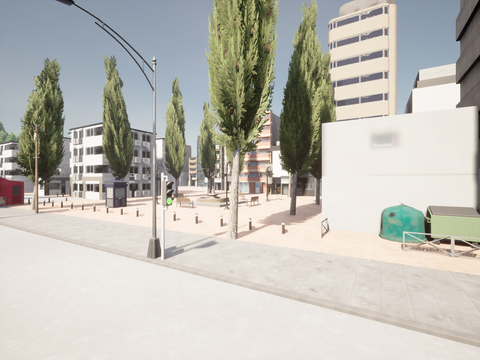}
    \caption{Town02}
\end{subfigure}
\hfill
\begin{subfigure}[b]{0.24\textwidth}
    \includegraphics[width=\textwidth]{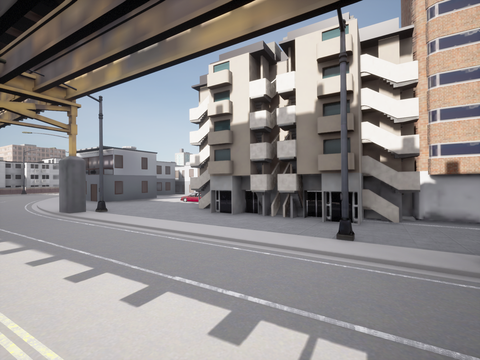}
    \caption{Town03}
\end{subfigure}
\hfill
\begin{subfigure}[b]{0.24\textwidth}
    \includegraphics[width=\textwidth]{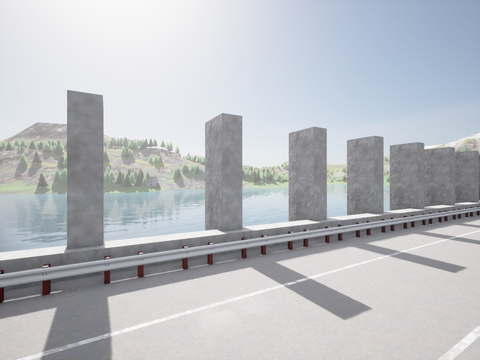}
    \caption{Town04}
\end{subfigure}

\vspace{0.1cm} 

\begin{subfigure}[b]{0.24\textwidth}
    \includegraphics[width=\textwidth]{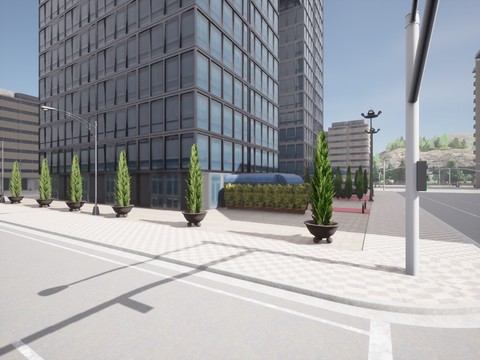}
    \caption{Town05}
\end{subfigure}
\hfill
\begin{subfigure}[b]{0.24\textwidth}
    \includegraphics[width=\textwidth]{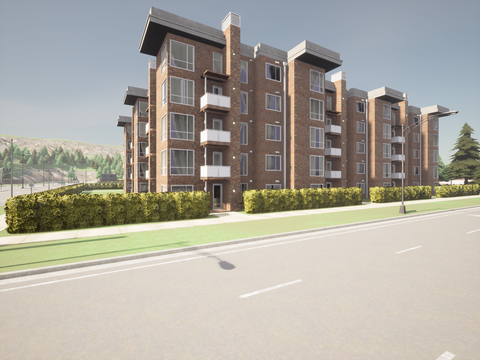}
    \caption{Town06}
\end{subfigure}
\hfill
\begin{subfigure}[b]{0.24\textwidth}
    \includegraphics[width=\textwidth]{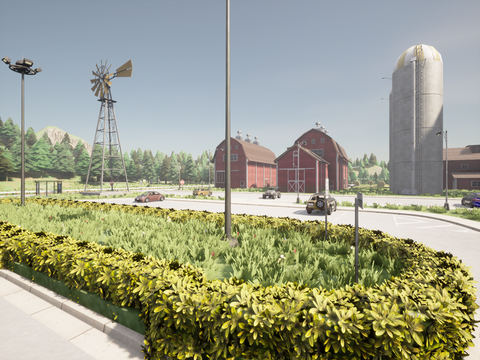}
    \caption{Town07}
\end{subfigure}
\hfill
\begin{subfigure}[b]{0.24\textwidth}
    \includegraphics[width=\textwidth]{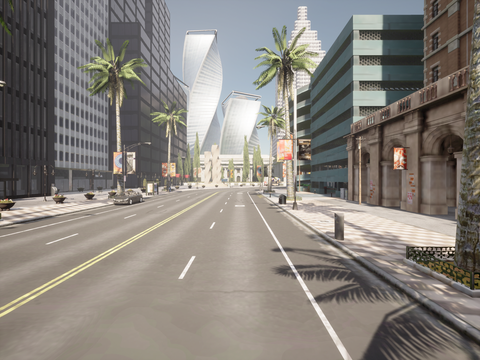}
    \caption{Town08}
\end{subfigure}\vspace{-1em}

\caption{Representative views of 8 different towns.}
\label{fig:towns_views}
\end{figure}

GameIR was collected using CARLA \cite{CARLA}, an autonomous driving simulator developed based on the UE4 game engine. CARLA provided 8 towns: Town01, Town02, Town03, Town04, Town05, Town06, Town07, and Town08. Each town has a distinct style and environment, including various simulation entities such as weather, roads, buildings, vehicles, pedestrians, and vegetation. Fig.~\ref{fig:towns_views} gives example views of these towns. For each town, we collected two types of scenes: the static autonomous driving scene where there were no other moving vehicles in the scene; and the dynamic autonomous driving scene, where there were other moving vehicles in the scene. Data were collected by controlling an agent vehicle driving in different towns with different camera setups. There were many spawn points for driving agent vehicles in each town, and after initializing the agent vehicle, we set it to autonomous driving mode. Different cameras were initialized and attached to the agent vehicle, which dynamically recorded the surrounding data as the vehicle drove through. During data collection, we set the CARLA simulation to synchronous mode, which prevented discrepancies between the camera capture and storage to avoid frame drops in the stored camera photos.\vspace{-.5em}


\vspace{-3em}


\begin{figure}[htbp]
    \centering
    \begin{tabular}{
        >{\centering\arraybackslash}m{0.05\textwidth} 
        >{\centering\arraybackslash}m{0.3\textwidth}
        >{\centering\arraybackslash}m{0.3\textwidth}
        >{\centering\arraybackslash}m{0.3\textwidth}
    }
    & \textbf{RGB} & \textbf{Segmentation Map} & \textbf{Depth map} \\
    \rotatebox{90}{\textbf{LR}} & 
    \includegraphics[width=0.3\textwidth]{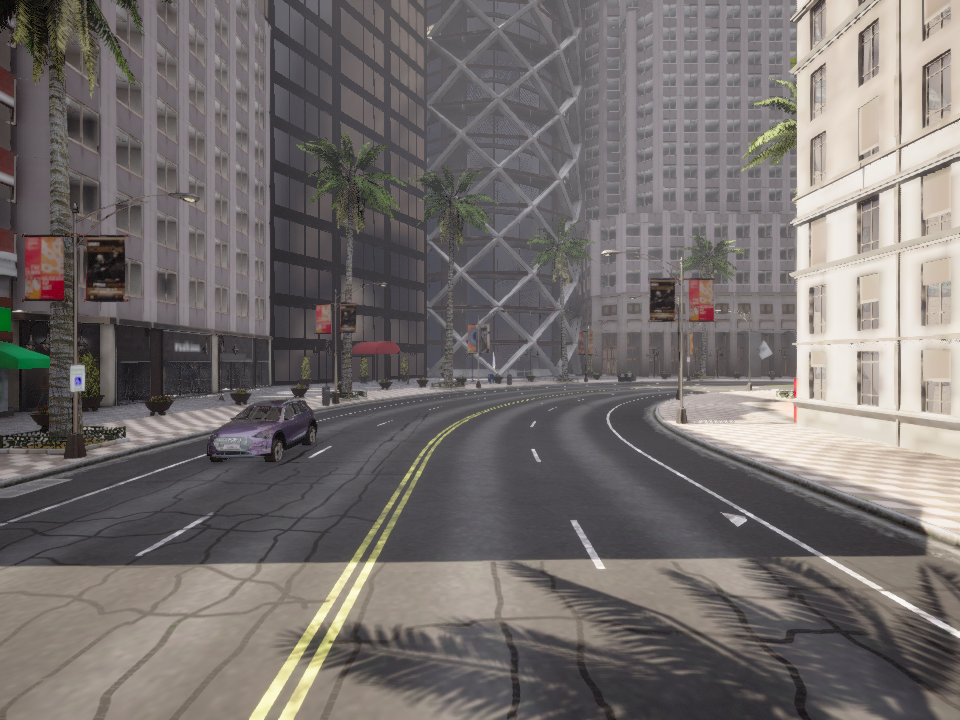} & 
    \includegraphics[width=0.3\textwidth]{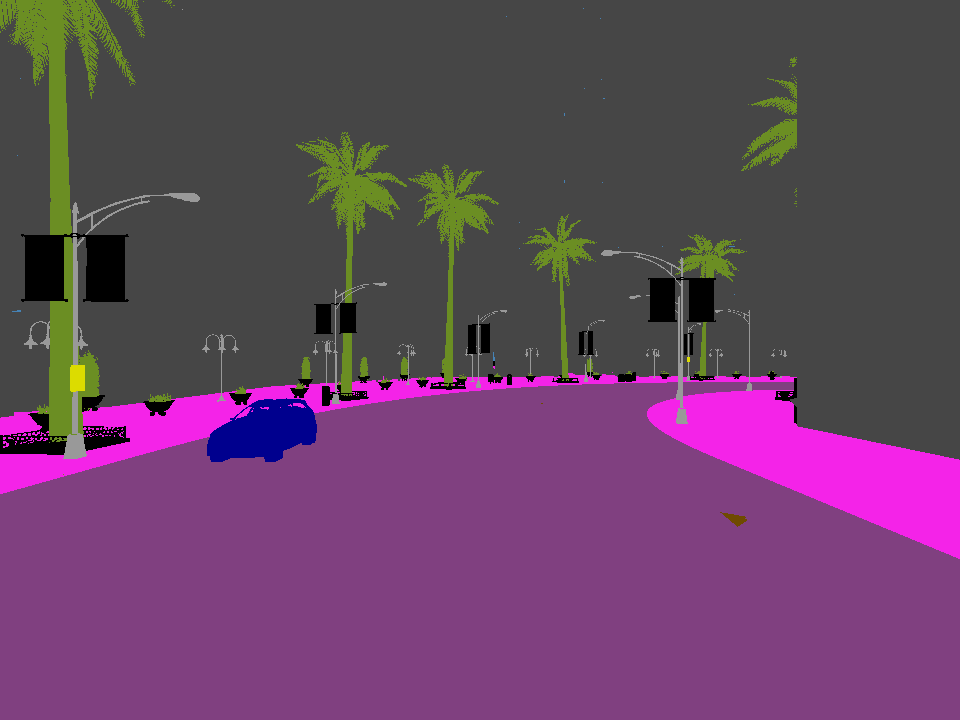} & 
    \includegraphics[width=0.3\textwidth]{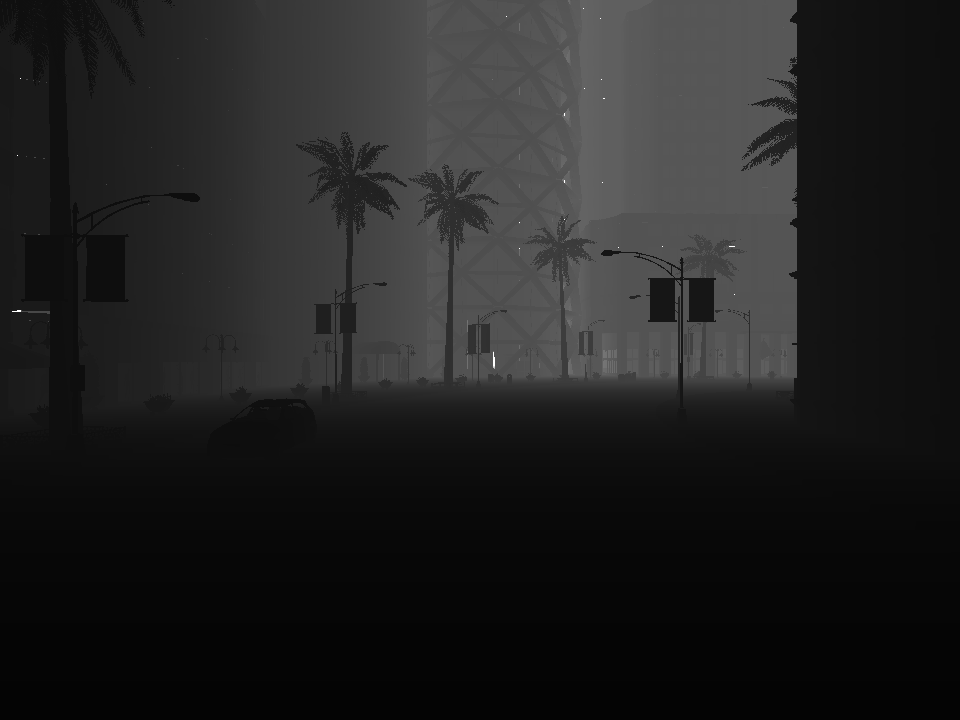} \\
    \rotatebox{90}{\textbf{HR}} & 
    \includegraphics[width=0.3\textwidth]{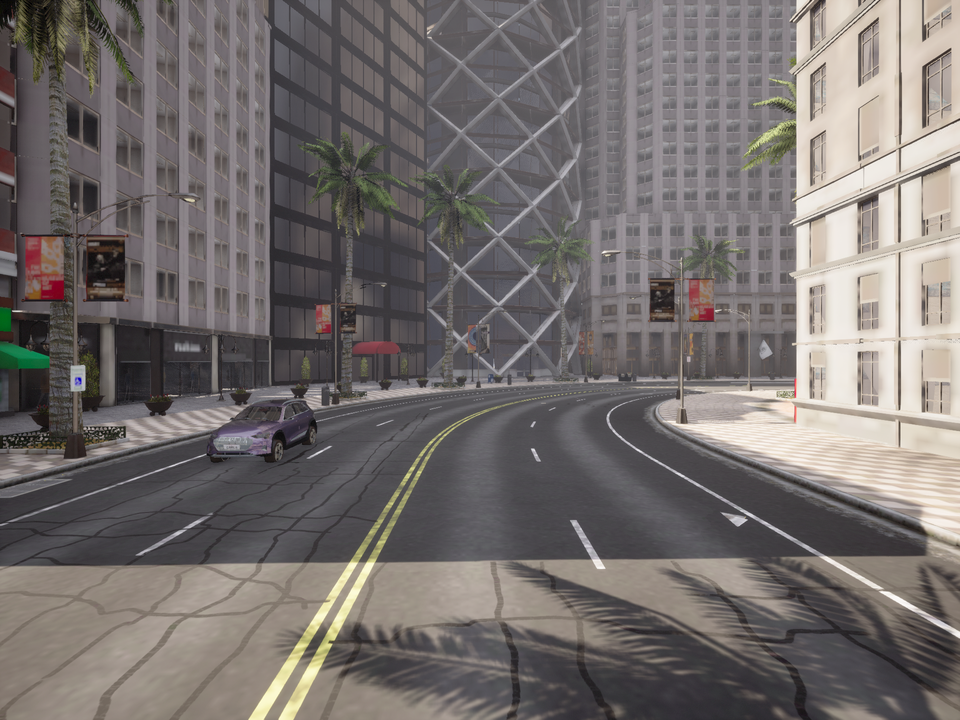} & 
    \includegraphics[width=0.3\textwidth]{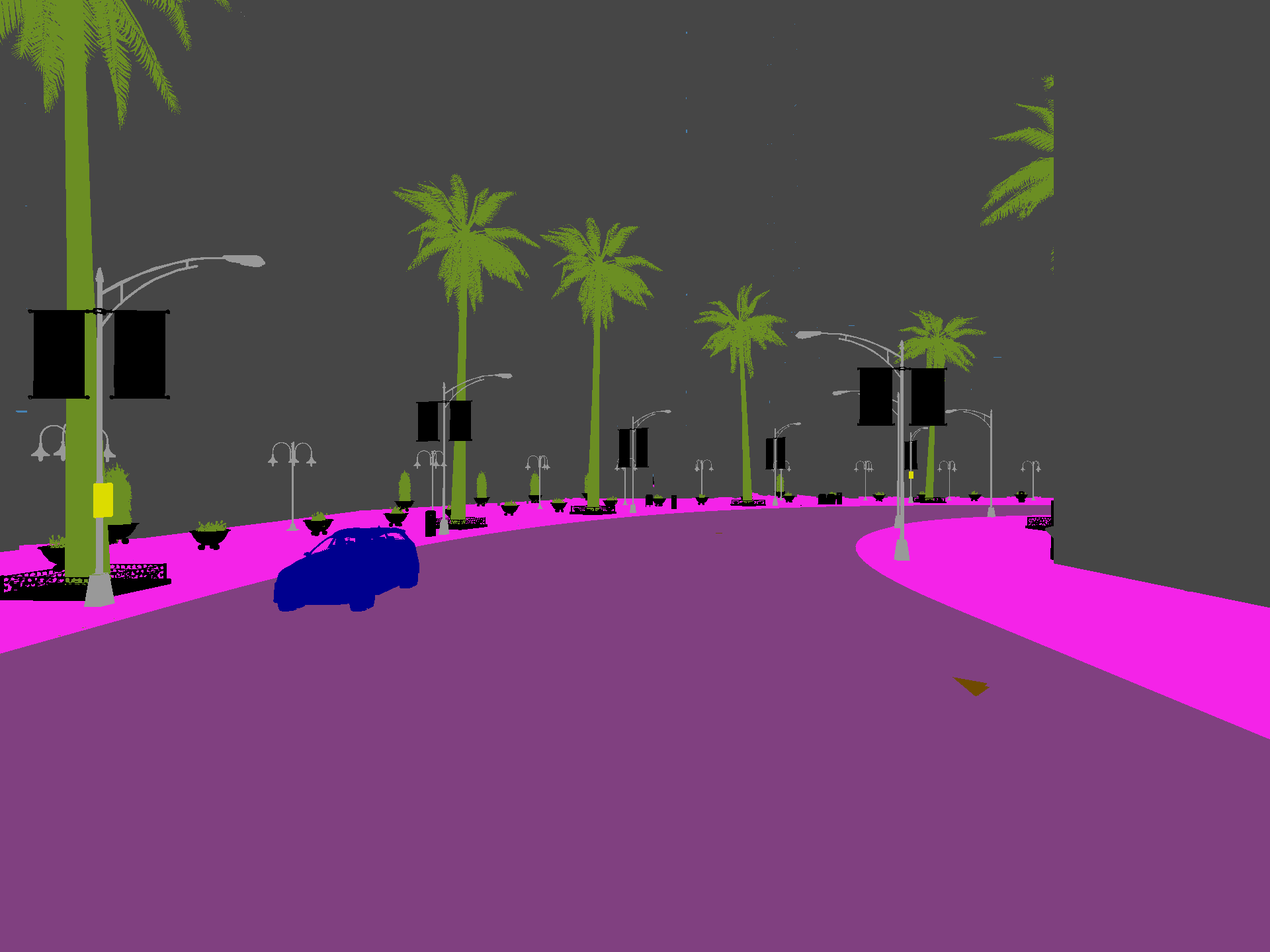} & 
    \includegraphics[width=0.3\textwidth]{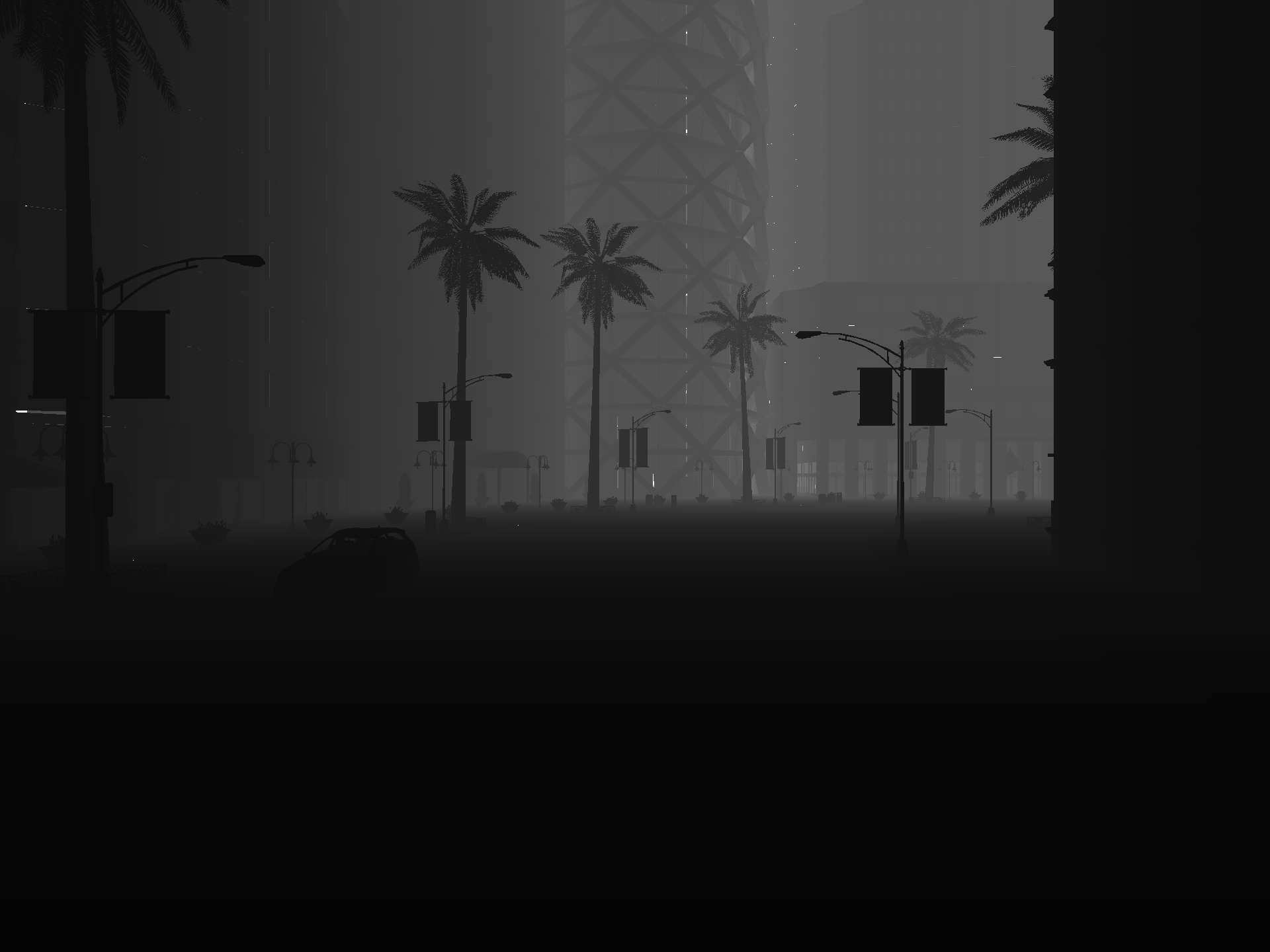} \\
    \end{tabular}\vspace{-1em}
    \caption{An example of GameIR-SR dataset, consisting of paired LR-HR (LR top row, HR bottom row) RGB images with associated GBuffer information.}
    \label{fig:sr_dataset_example}
\end{figure}\vspace{-2em}

\subsection{GameIR-SR: Dataset for Super-Resolution}

For both static and dynamic autonomous driving scenes, we randomly selected 20 spawn points in each of the 8 towns, totaling 320 scenes for the GameIR-SR dataset. To provide ground truth for super-resolution, we placed one set of HR cameras and one set of LR cameras at the front of the agent vehicle, each set capturing synchronized RGB images, segmentation maps, and depth maps with 1920x1440 resolution and 960x720 resolution, respectively.

\begin{figure}[htbp]
\centering

\begin{subfigure}[b]{0.49\textwidth}
    \includegraphics[width=\textwidth]{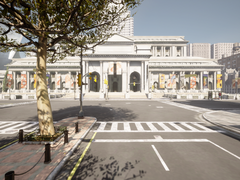}
    \caption{True LR}
\end{subfigure}
\hfill 
\begin{subfigure}[b]{0.49\textwidth}
    \includegraphics[width=\textwidth]{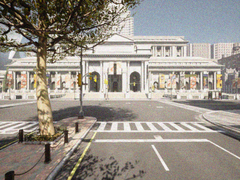}
    \caption{Pseudo LR}
\end{subfigure}\vspace{-1em}

\caption{True LR input versus pseudo LR input. The pseudo LR input is downsampled from the HR input with added noise and blur as commonly used degradation\cite{zhang2017learning, zhang2018residual}.}
\label{fig:gaming_characteristics}
\end{figure}

Each video in the GameIR-SR dataset is 2-second long at 30fps, totaling 60 frames. During capture, the GBuffer data from the deferred rendering phase, as well as the cameras's intrinsic parameters and extrinsic 6-DoF parameters, were also collected synchronously. Fig.~\ref{fig:sr_dataset_example} gives an example of the GameIR-SR dataset. Finally, GameIR-SR has 19200 LR-HR paired ground-truth frames from 320 LR-HR paired videos, along with the corresponding GBuffers and camera parameters. The ground-truth LR frames are clear and sharp, different from the pseudo LR images generated by degrading the HR images, as illustrated in Fig.~\ref{fig:gaming_characteristics}. Such ground-truth LR-HR pairs can better serve as training data for super-resolution methods targeting at gaming content, where the real degradation features can be learned to improve the models' performance.\vspace{-1em}


\vspace{-2em}

\subsection{GameIR-NVS: Dataset for Novel View Synthesis}

\begin{figure}[htbp]
    \centering
    \begin{tabular}{
        >{\centering\arraybackslash}m{0.05\textwidth} 
        >{\centering\arraybackslash}m{0.3\textwidth}
        >{\centering\arraybackslash}m{0.3\textwidth}
        >{\centering\arraybackslash}m{0.3\textwidth}
    }
    & \textbf{RGB} & \textbf{Segmentation Map} & \textbf{Depth map} \\
    \rotatebox{90}{\textbf{Front}} & 
    \includegraphics[width=0.28\textwidth]{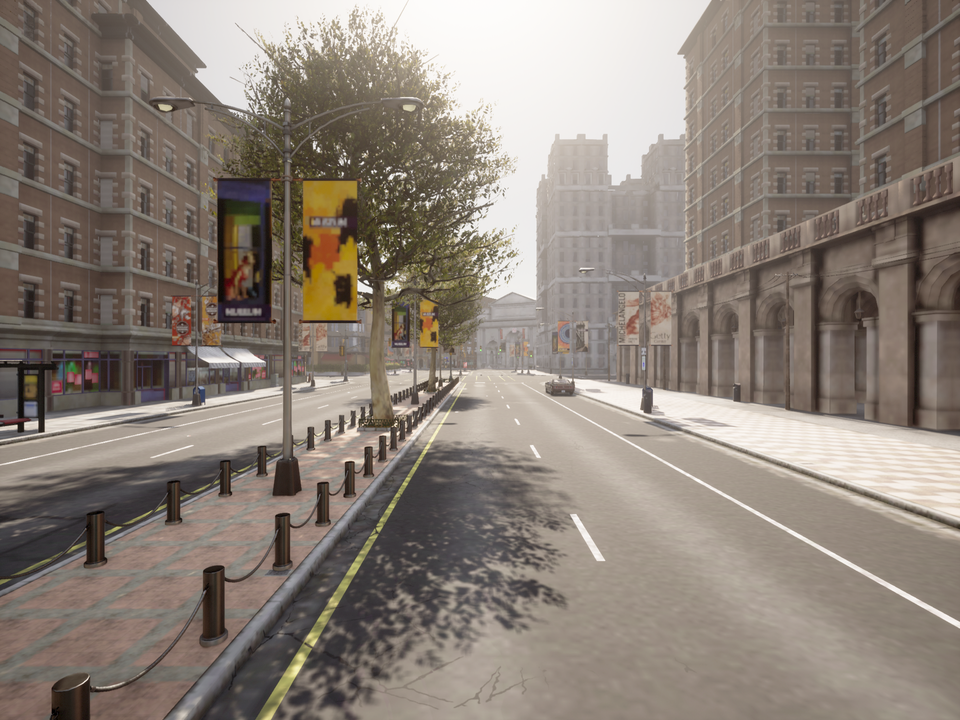} & 
    \includegraphics[width=0.28\textwidth]{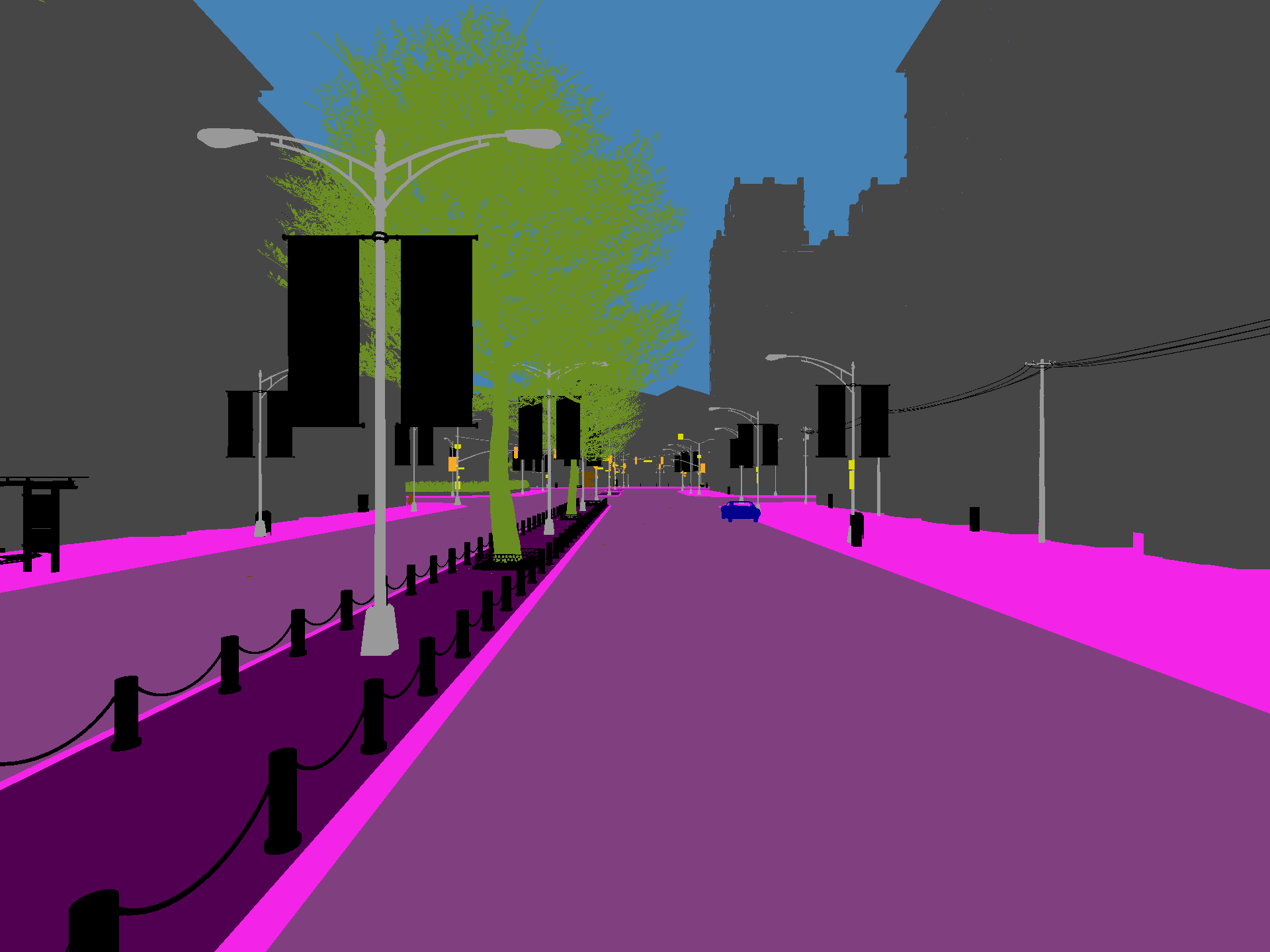} & 
    \includegraphics[width=0.28\textwidth]{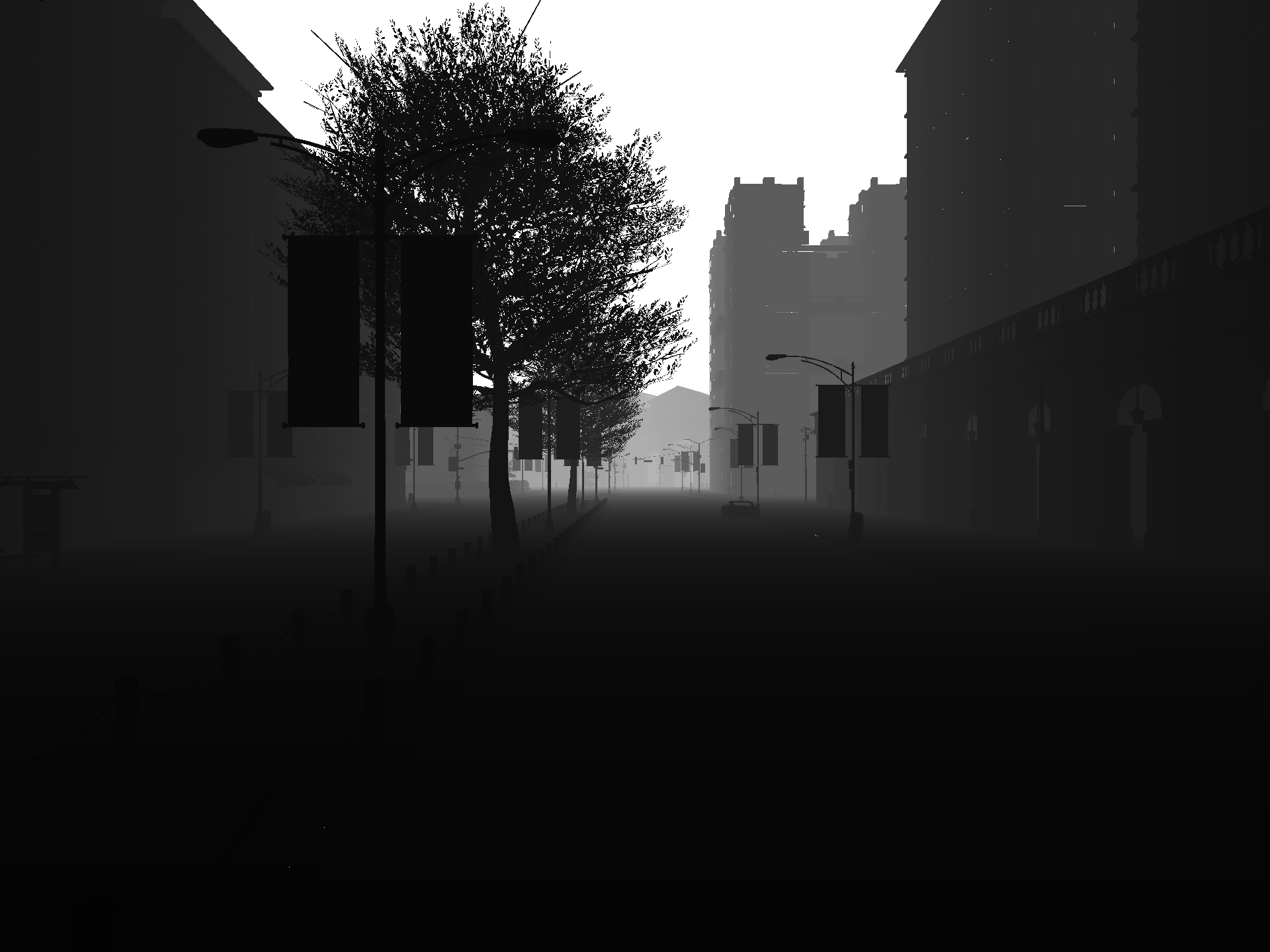} \\
    \rotatebox{90}{\textbf{Left 60°}} & 
    \includegraphics[width=0.28\textwidth]{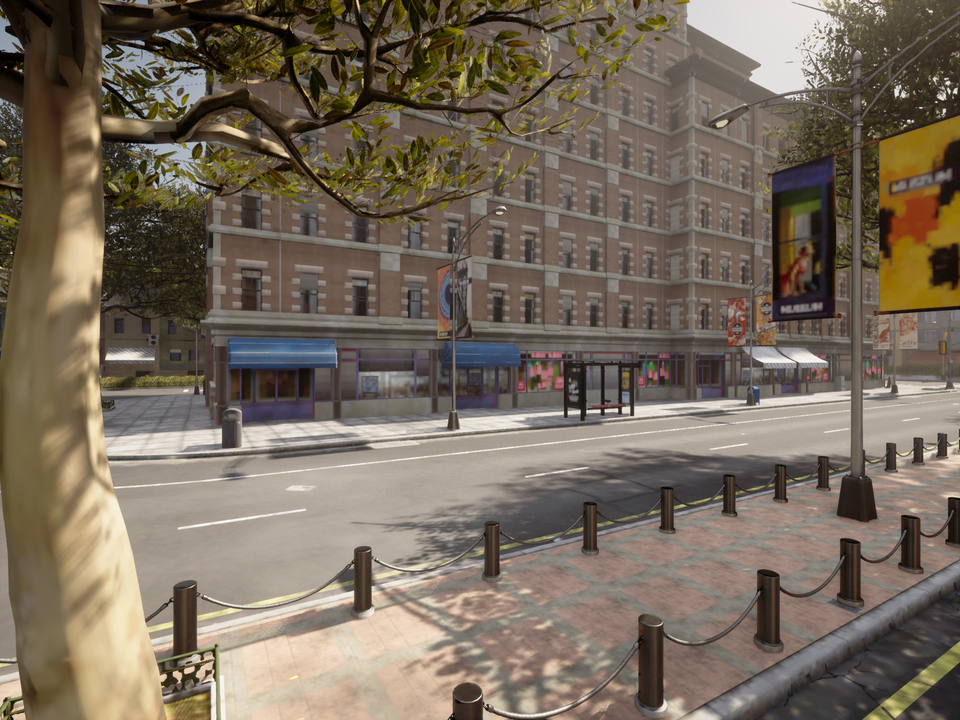} & 
    \includegraphics[width=0.28\textwidth]{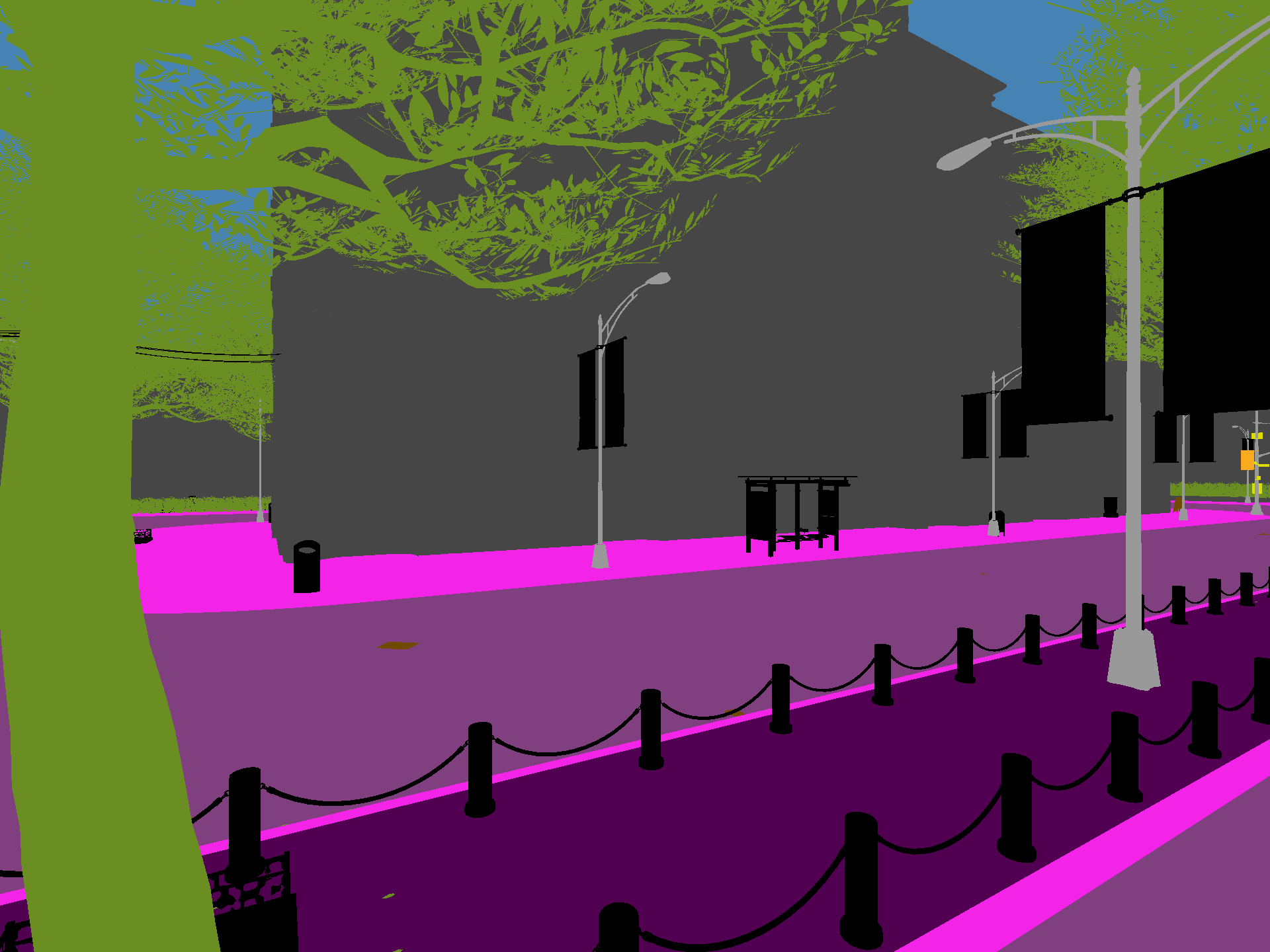} & 
    \includegraphics[width=0.28\textwidth]{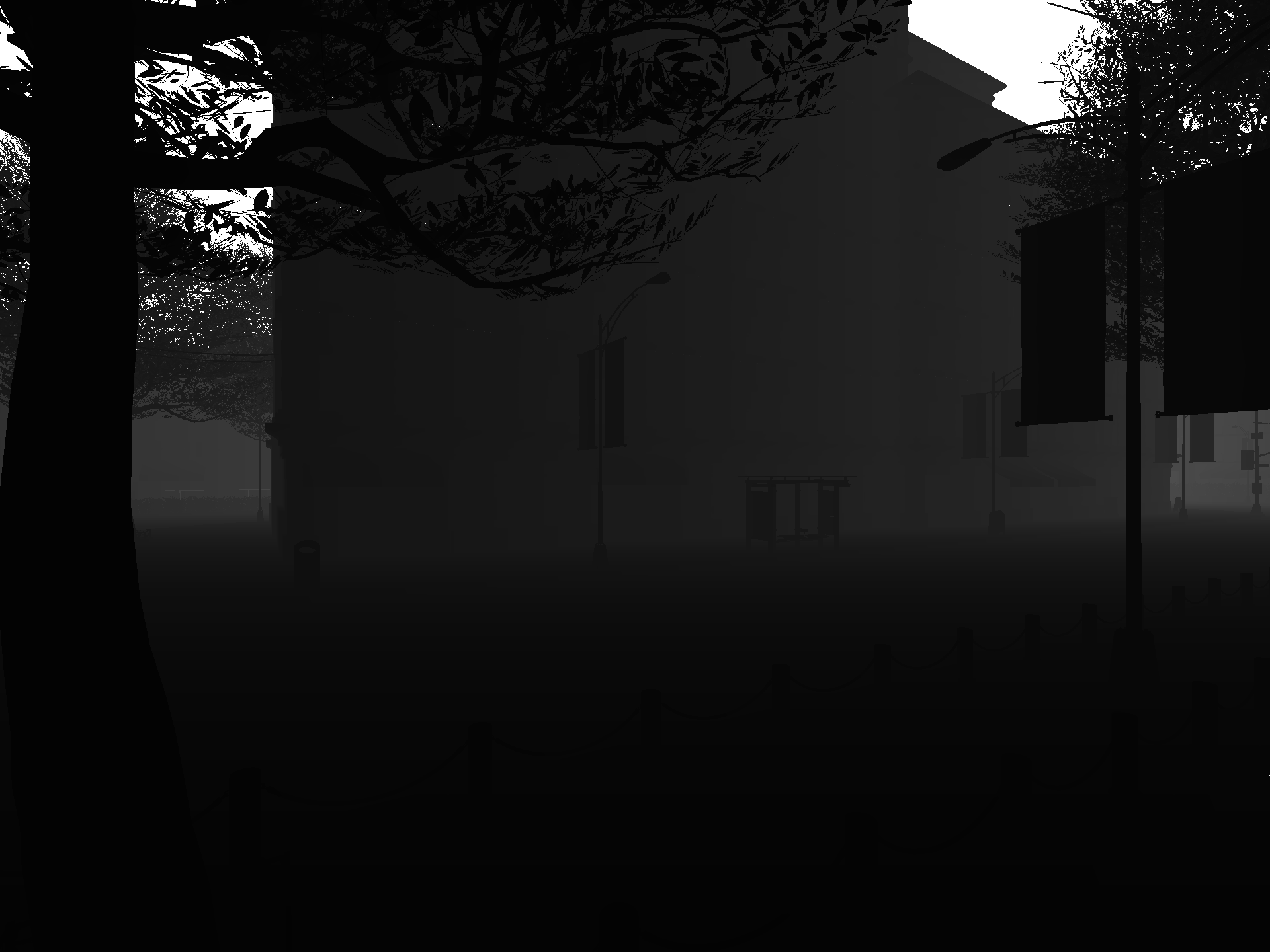} \\
    \rotatebox{90}{\textbf{Left 120°}} & 
    \includegraphics[width=0.28\textwidth]{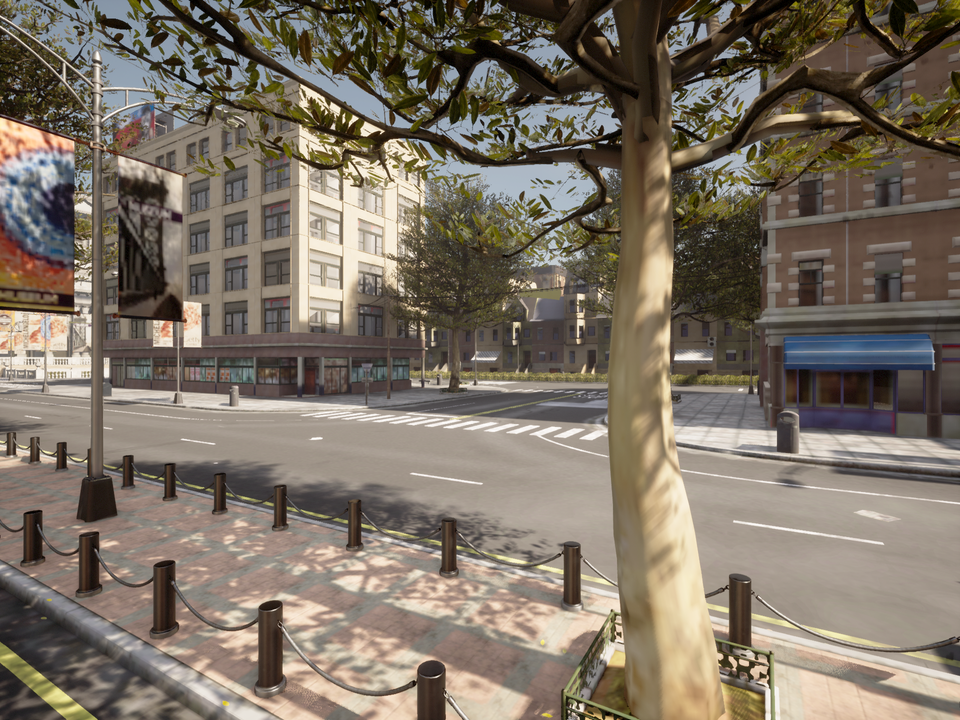} & 
    \includegraphics[width=0.28\textwidth]{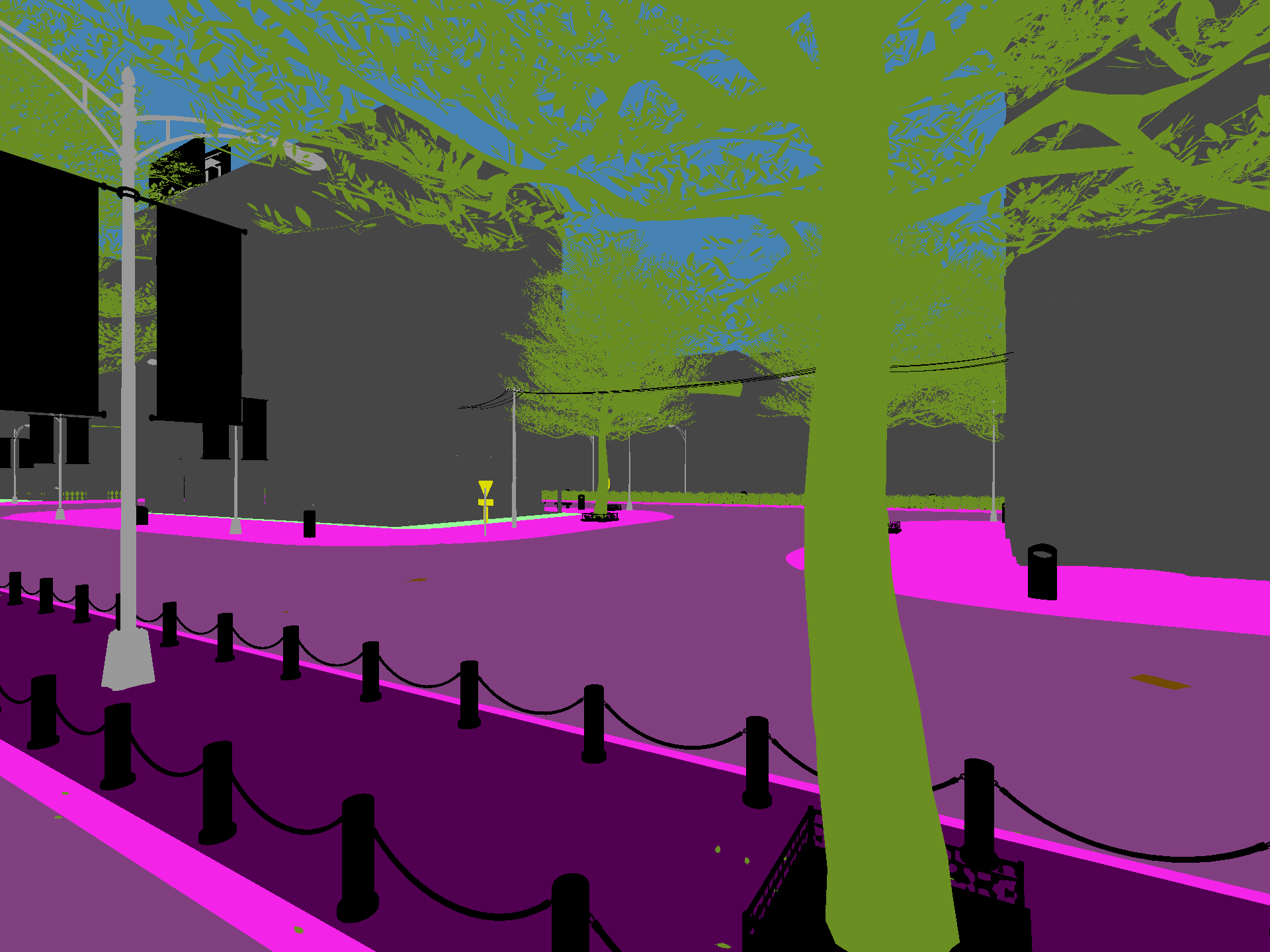} & 
    \includegraphics[width=0.28\textwidth]{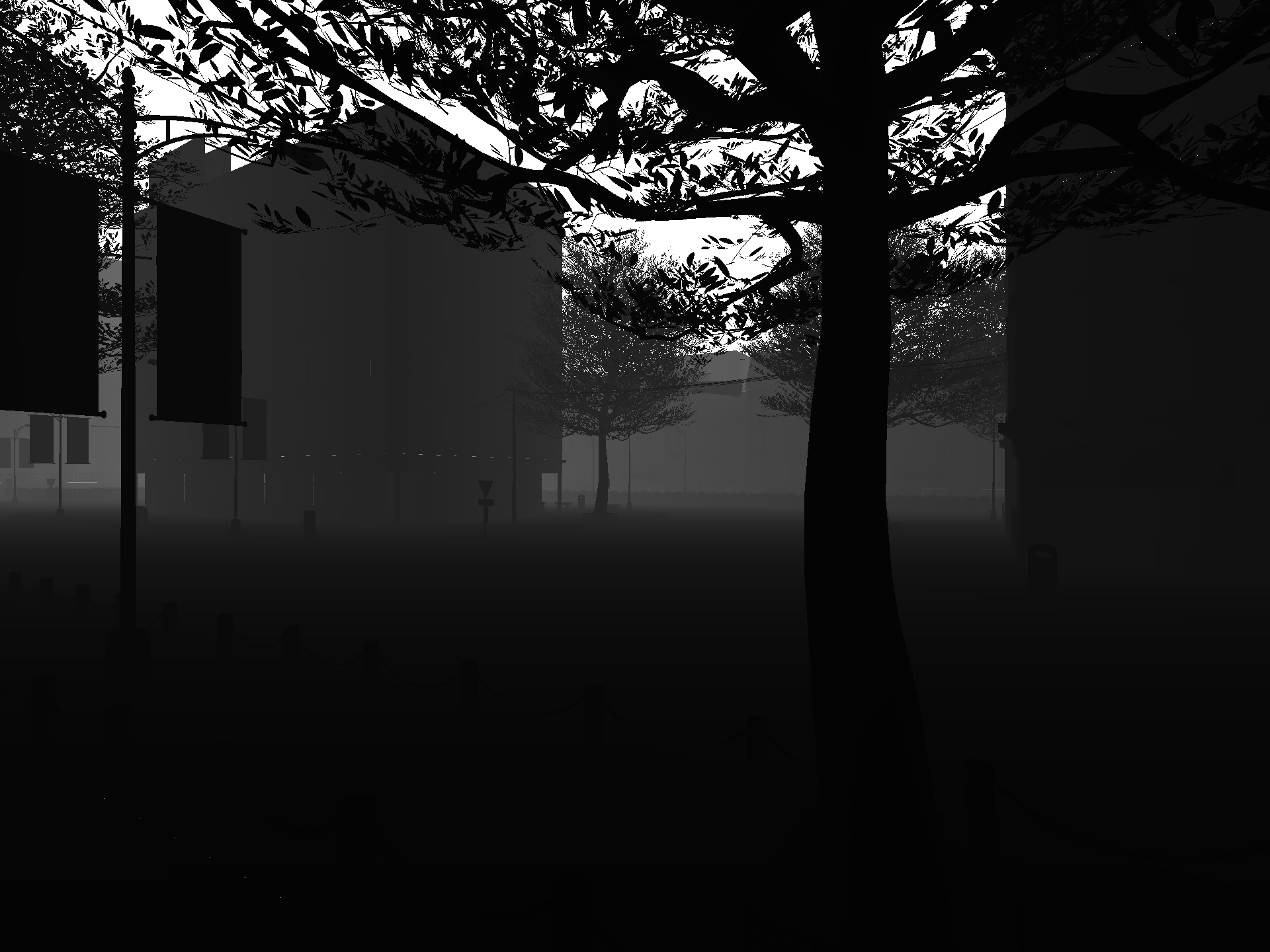} \\
    \rotatebox{90}{\textbf{Back}} & 
    \includegraphics[width=0.28\textwidth]{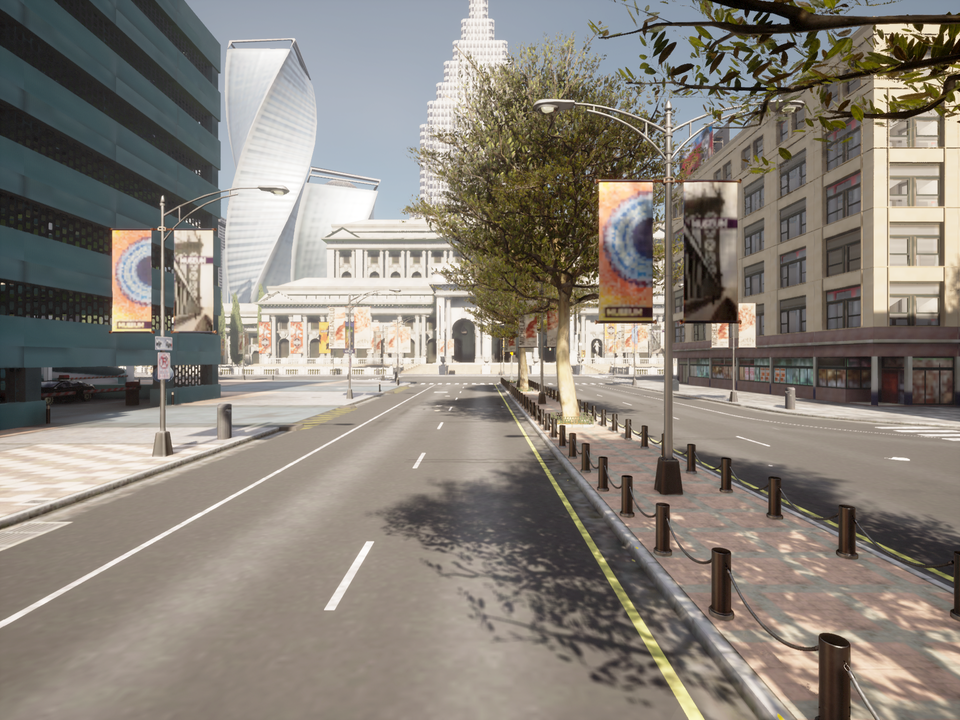} & 
    \includegraphics[width=0.28\textwidth]{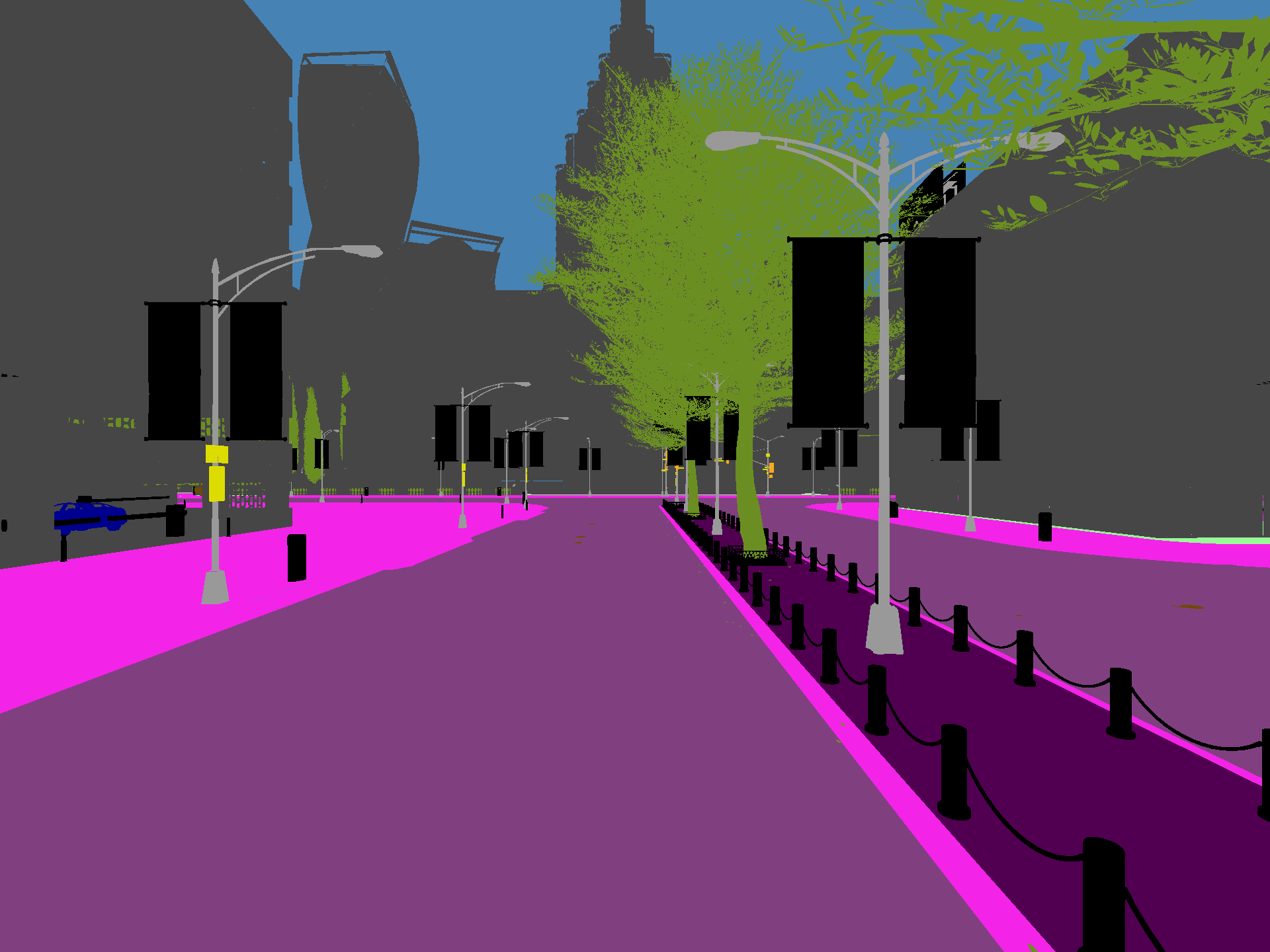} & 
    \includegraphics[width=0.28\textwidth]{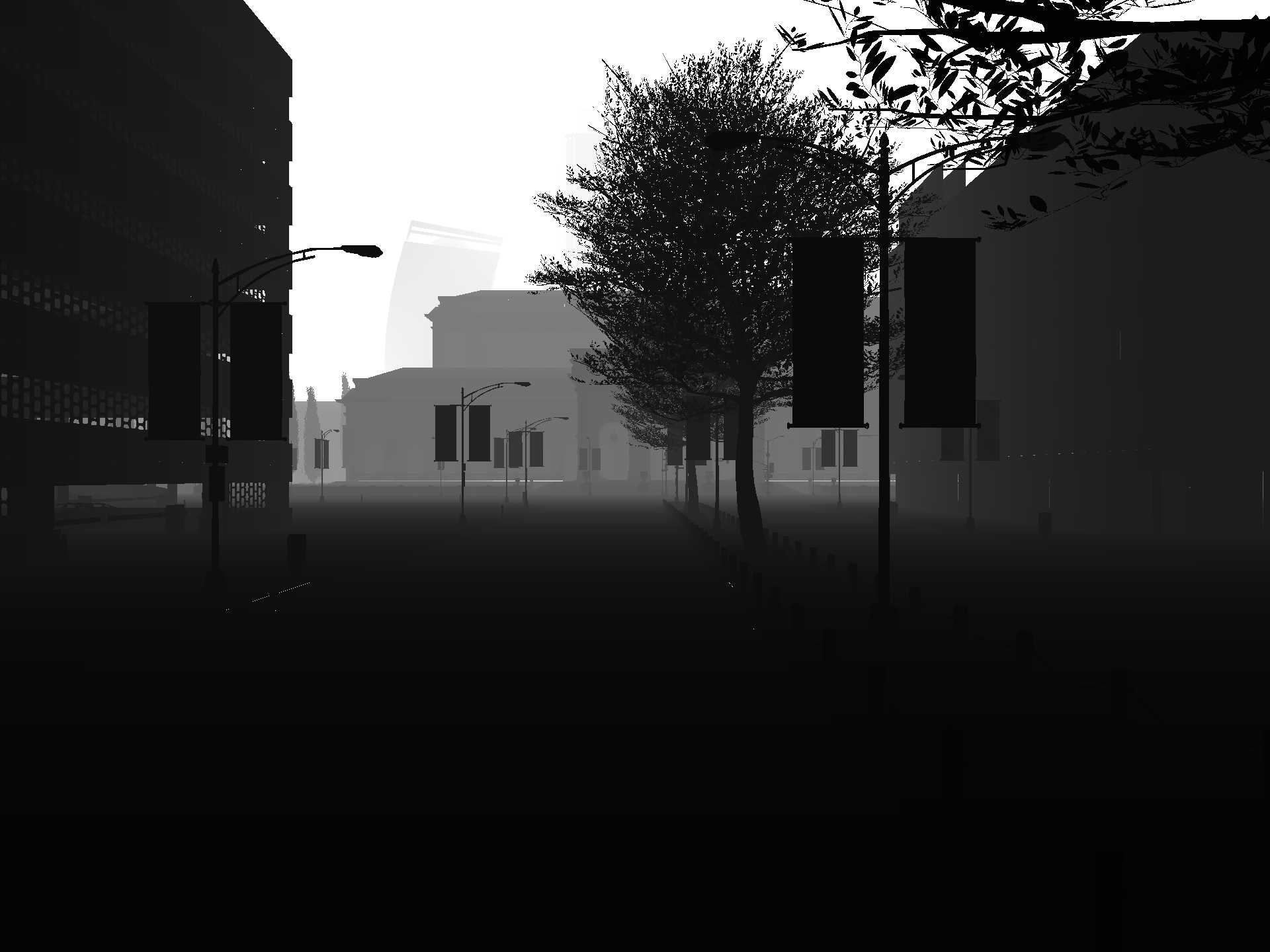} \\
    \rotatebox{90}{\textbf{Right 120°}} & 
    \includegraphics[width=0.28\textwidth]{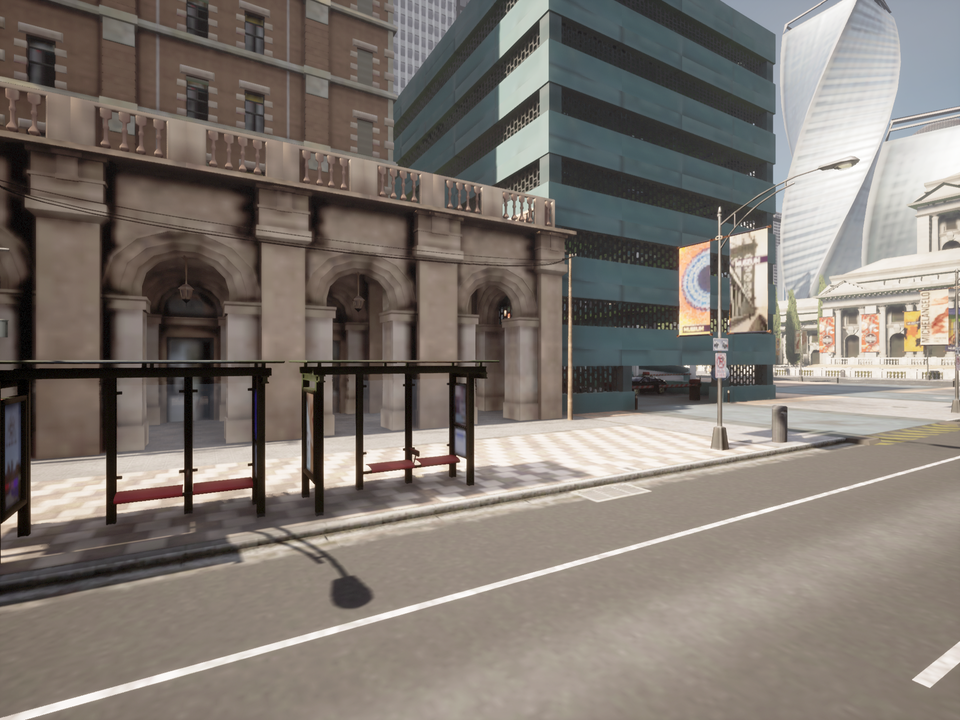} & 
    \includegraphics[width=0.28\textwidth]{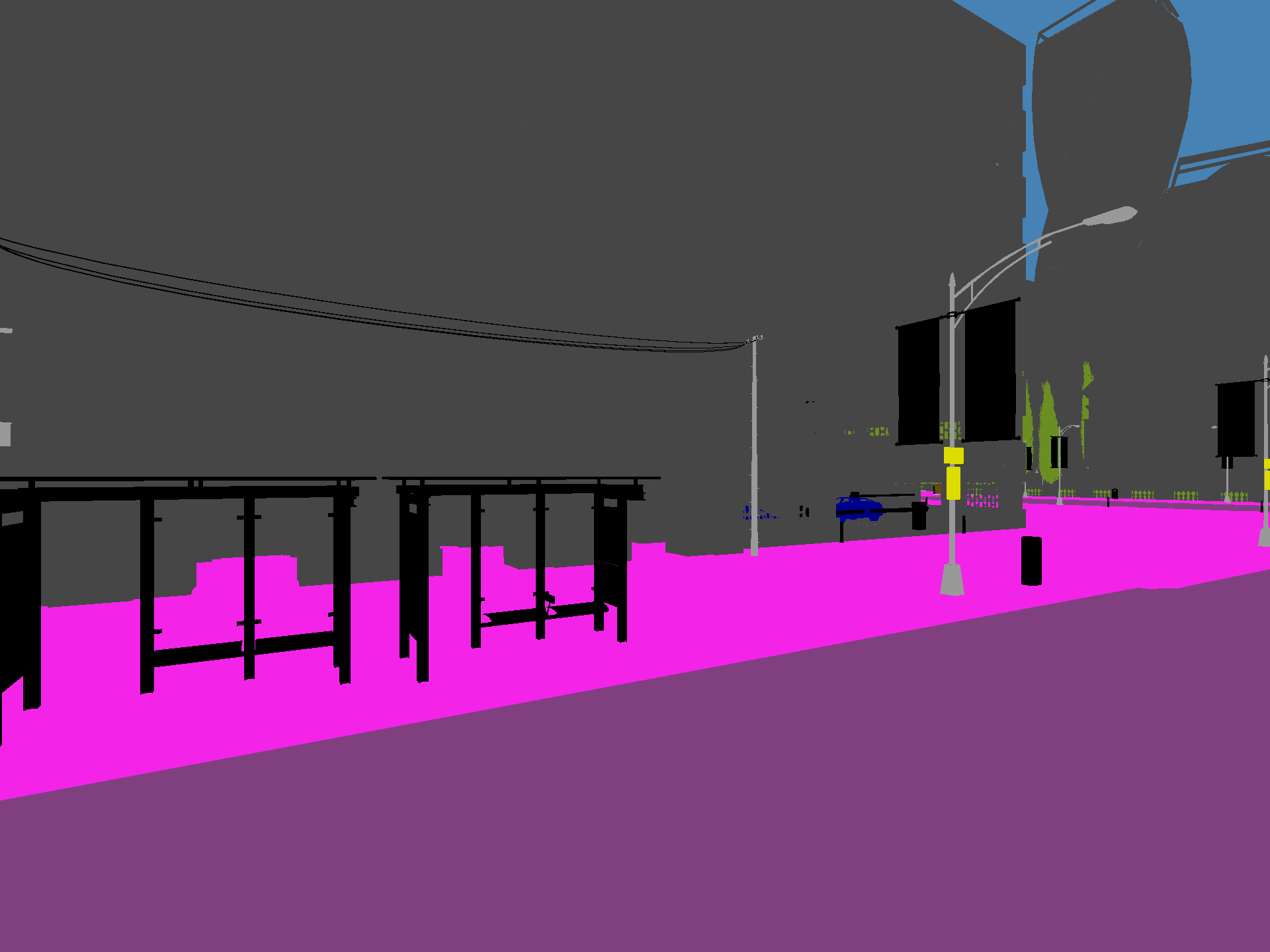} & 
    \includegraphics[width=0.28\textwidth]{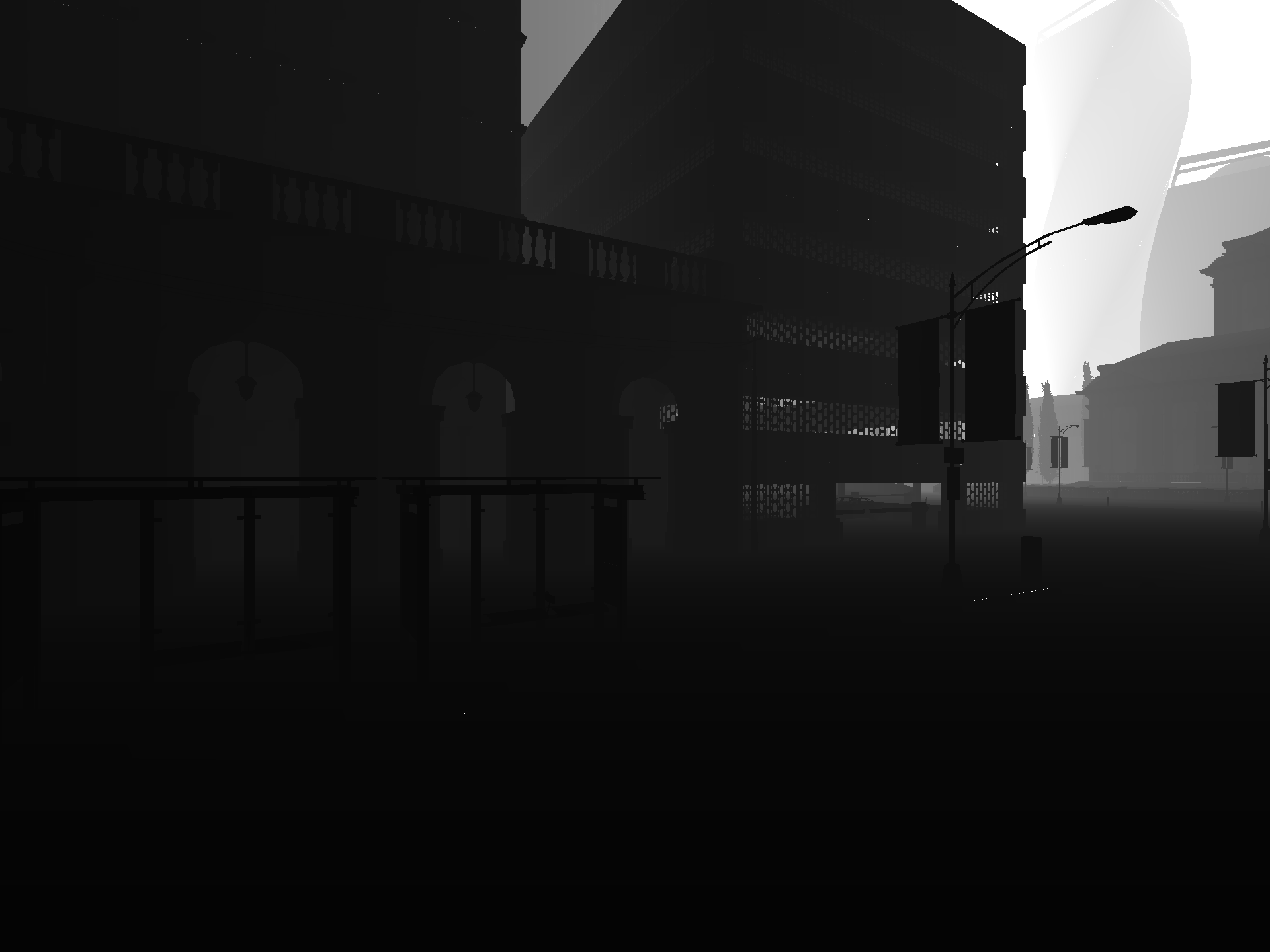} \\
    \rotatebox{90}{\textbf{Right 60°}} & 
    \includegraphics[width=0.28\textwidth]{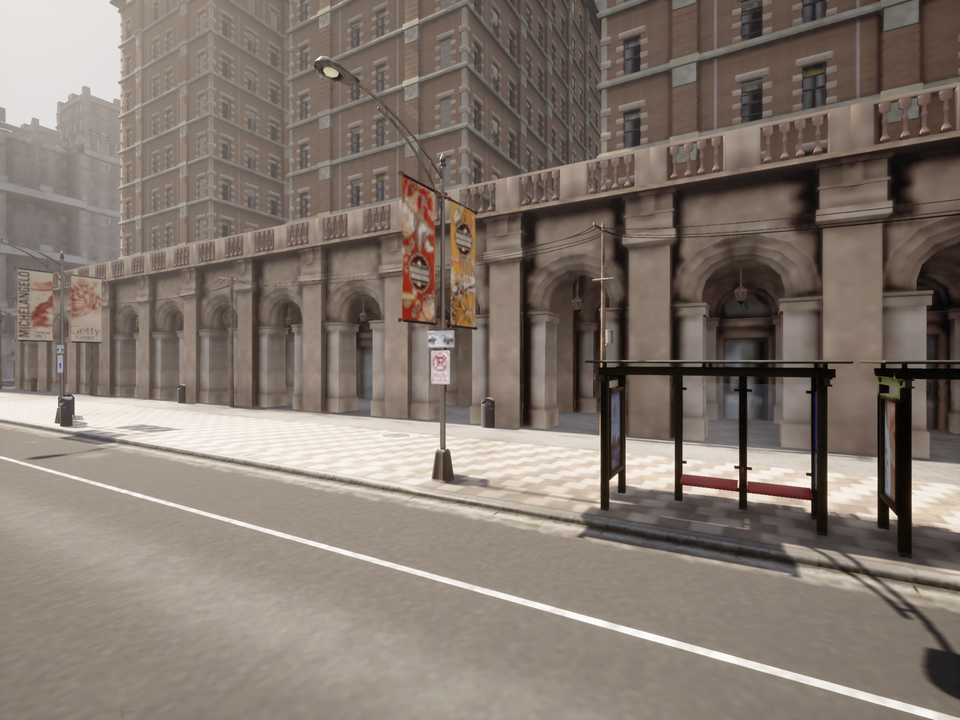} & 
    \includegraphics[width=0.28\textwidth]{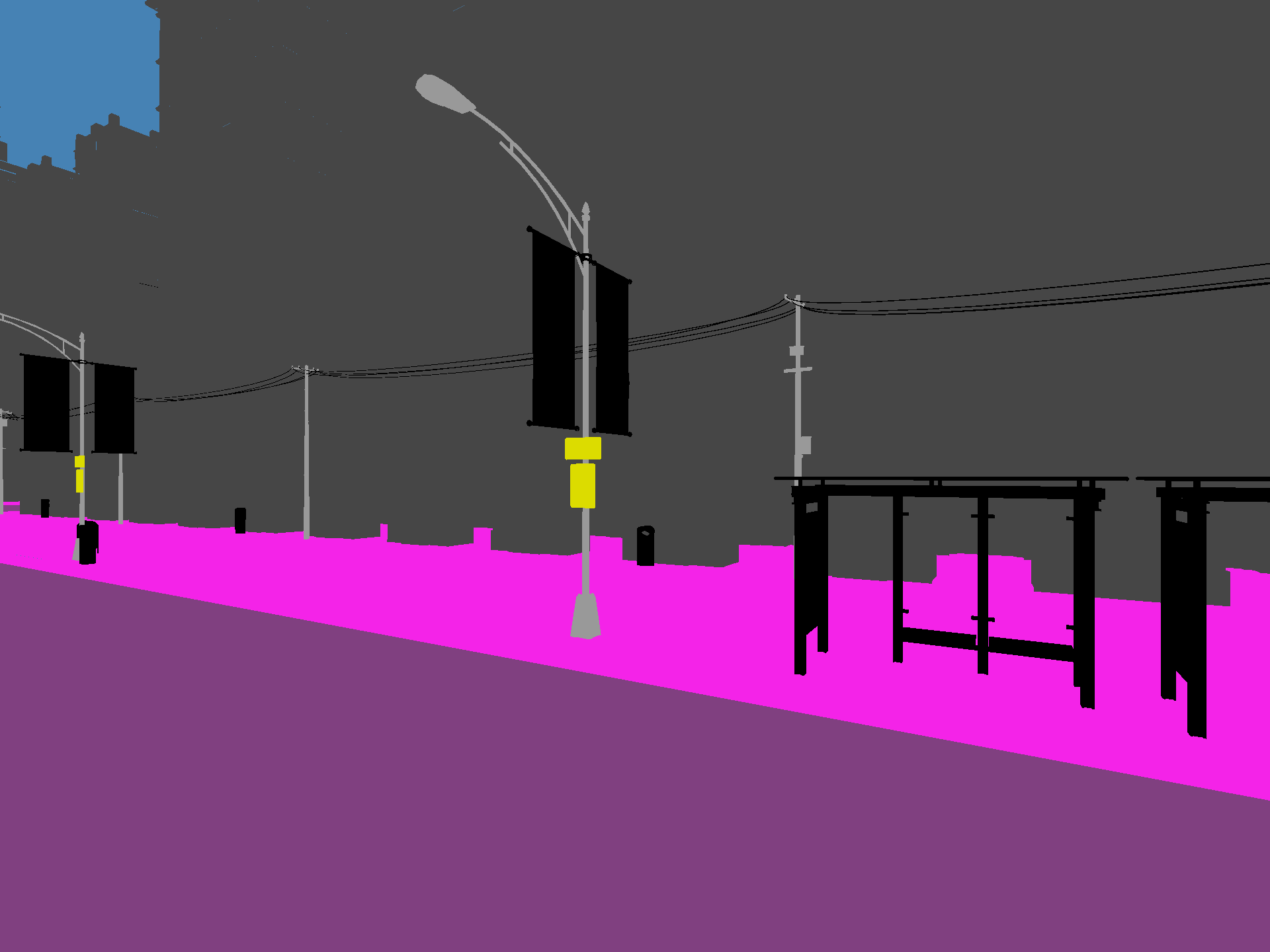} & 
    \includegraphics[width=0.28\textwidth]{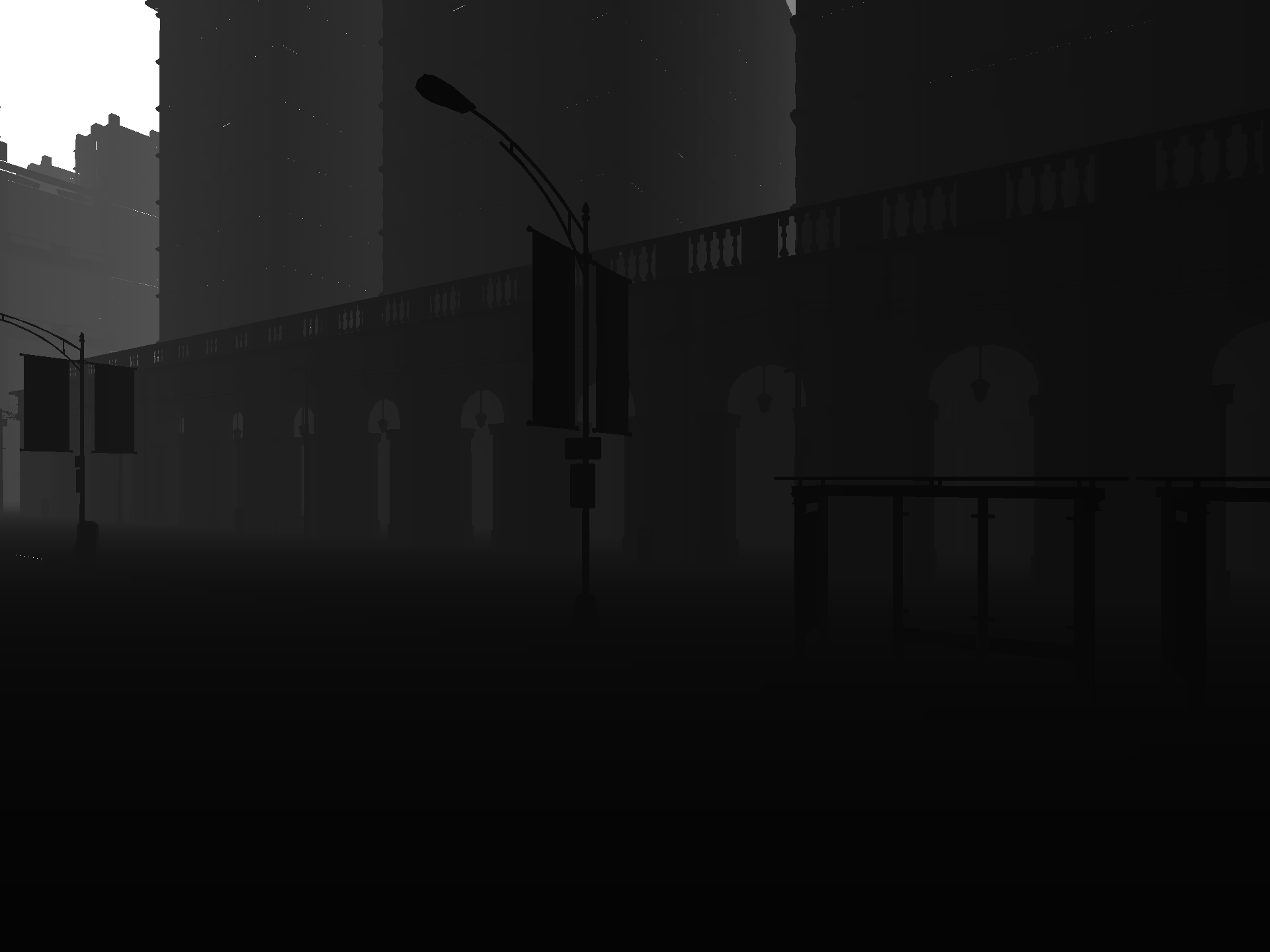} \\
    \end{tabular}\vspace{-1em}
    \caption{An example of our GameIR-NVS dataset, consisting of 6 views of RGB images with associated segmentation maps and depth maps. From the top to bottom, the sequence gives views of a 360° rotation, starting from the front.}
    \label{fig:multiview_dataset_example}
\end{figure}

For NVS, we only collected data with static autonomous driving scenes. We randomly selected 20 spawn points in each of the 8 towns and recorded 160 scenes in total for the GameIR-NVS dataset. To provide ground truth for NVS, we placed 6 sets of cameras in 6 directions around the agent vehicle: front view, left 60° view, right 60° view, left 120° view, right 120° view, and back view. Each set captured the RGB images, semantic segmentation maps, and depth maps at the resolution of 1920x1440 when the vehicle drove through different parts of the towns. Adjacent cameras have some overlapping field-of-view. For each scene, the video is 2-second long at 30fps, totaling 60 frames. Fig.~\ref{fig:multiview_dataset_example} gives an example of the GameIR-NVS dataset. The camera intrinsic parameters and the 6-DoF camera extrinsic parameters for each frame are also recorded. Finally, the GameIR-NVS dataset comprises 960 videos from 160 scenes, totaling 57,600 HR frames. These 360-degree scene-level synthetic data are suitable for training and evaluating NVS methods over gaming content. In addition, the associated depth maps and segmentation maps can be leveraged by NVS algorithms to further improve the generation performance.
\vspace{-1em}

\section{Evaluation of Super-Resolution over GameIR-SR}\vspace{-.5em}

Our evaluation has 3 progressive stages: test pretrained SISR models, test finetuned models using the GameIR-SR training set, and test modified SISR models using the GBuffer information as additional inputs or as generative conditions, trained over the GameIR-SR training set. \vspace{.3em}

\noindent
\textbf{Tested SISR Methods.} We evaluated 3 methods: Anime4K\cite{Anime4K} that is specially designed and trained for anime/cartoon images, Real-ESRGAN\cite{Realesrgan} that handles diverse real-world image degradations, and Adacode\cite{AdaCode} that employs a learned codebook-based visual representation as learned generative priors. These methods have their unique strengths in different application areas and give SOTA performance over both anime/cartoon and real-world images. By evaluating their pretrained, finetuned, and modified improved models, we can obtain a good assessment of how current SISR methods perform over gaming content.  \vspace{.3em}

\noindent\textbf{Implementation Details.} We used data from Town01-04 and Town06-08 for training, and data from Town05 for testing. This ensures that training and test data have distinct styles and town structures. We used published source code by Anime4K\cite{Anime4K}, Real-ESRGAN\cite{Realesrgan}, and Adacode\cite{AdaCode}. During training or finetuning, we followed the original methodologies and hyperparameters described in each method's seminal papers. We used eight V100 GPUs for training and a single V100 GPU for testing.\vspace{.3em}

\noindent
\textbf{Evaluation Metrics.} We evaluated PSNR, SSIM, FID, and LPIPS. PSNR and SSIM focus on pixel-level distortions. LPIPS highlights local image quality. FID measures the distance between the distributions of generated and real images, reflecting the overall perceptual quality. These metrics are widely used for restoration tasks to provide complementary perspectives on image quality.\vspace{-1.5em}

\begin{table}[htbp]\scriptsize
\centering
\resizebox{1.0\textwidth}{\height}{%
\renewcommand\arraystretch{1.5}
\begin{tabular}{c ccc ccc ccc ccc}
\toprule
            & \multicolumn{3}{c}{PSNR$\uparrow$}    & \multicolumn{3}{c}{SSIM$\uparrow$}   & \multicolumn{3}{c}{FID$\downarrow$} & \multicolumn{3}{c}{LPIPS$\downarrow$}    \\ 
\midrule
            & PT & \makecell[c]{FT\\(pseudo)} & \makecell[c]{FT\\(GT)}      & PT & \makecell[c]{FT\\(pseudo)} & \makecell[c]{FT\\(GT)}     & PT & \makecell[c]{FT\\(pseudo)} & \makecell[c]{FT\\(GT)}     & PT & \makecell[c]{FT\\(pseudo)} & \makecell[c]{FT\\(GT)}\\
Anime4k     & 30.9274    & 30.5923   & \textbf{31.2974} & 0.9008   & 0.8874 & \textbf{0.9057} & 4.7365  & 6.7408  & \textbf{4.6172}    & 0.0866  & 0.1069  & \textbf{0.0788}  \\
AdaCode     & 28.6791    & 27.9637   & \textbf{29.5762} & 0.8382   & 0.8146 & \textbf{0.8741} & 14.8789  & 17.6222  & \textbf{11.4608} & 0.0884  & 0.0899  & \textbf{0.0451}  \\
Real-ESRGAN & 29.1421    & 29.2743   & \textbf{30.2517} & 0.8639   & 0.8623 & \textbf{0.8916} & 17.7232  & 14.7961  & \textbf{8.3153}  & 0.0905  & 0.0706  & \textbf{0.0379}     \\ 
\bottomrule
\end{tabular}
}
\caption{Performance of pretrained (PT) and finetuned (FT) super-resolution methods, using pseudo-LR training data or ground-truth (GT) LR training data: PSNR and SSIM, the higher the better; LPIPS and FID, the lower the better.}
\label{tab:sr_pretrain_and_finetuned_results}
\end{table}\vspace{-4em}

\subsection{Performance without GBuffer}

\begin{figure}[htbp]
\centering
\begin{subfigure}[b]{0.24\textwidth}
    \includegraphics[width=\textwidth]{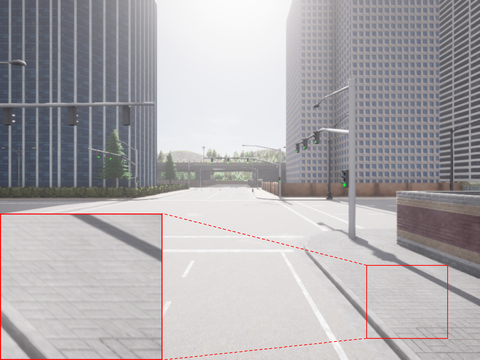}
    \captionsetup{labelformat=empty}
    \caption{LR}
\end{subfigure}
\hfill 
\begin{subfigure}[b]{0.24\textwidth}
    \includegraphics[width=\textwidth]{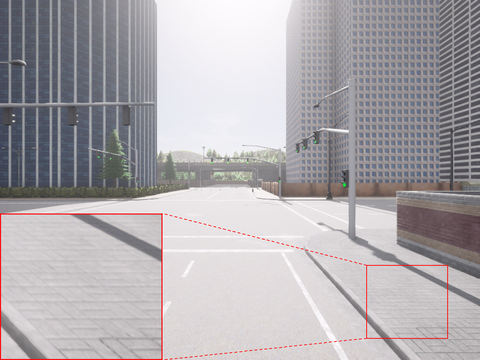}
    \captionsetup{labelformat=empty}
    \caption{\scriptsize Anime4K (PT)}
\end{subfigure}
\hfill
\begin{subfigure}[b]{0.24\textwidth}
    \includegraphics[width=\textwidth]{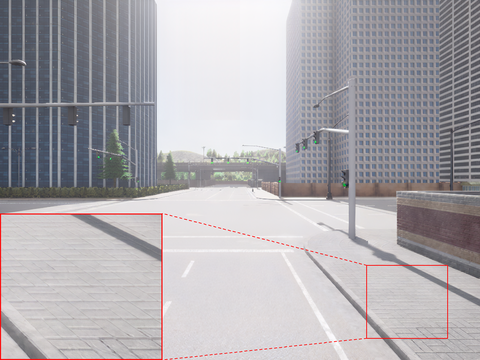}
    \captionsetup{labelformat=empty}
    \caption{AdaCode (PT)}
\end{subfigure}
\hfill
\begin{subfigure}[b]{0.24\textwidth}
    \includegraphics[width=\textwidth]{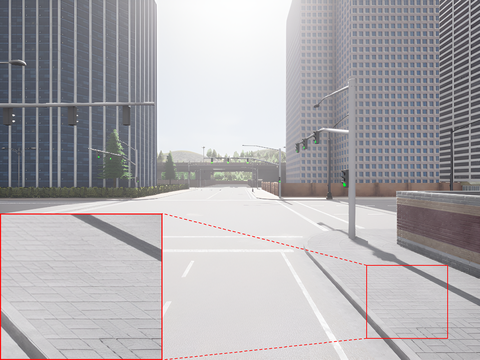}
    \captionsetup{labelformat=empty}
    \caption{Real-ESRGAN (PT)}
\end{subfigure}

\begin{subfigure}[b]{0.24\textwidth}
    \includegraphics[width=\textwidth]{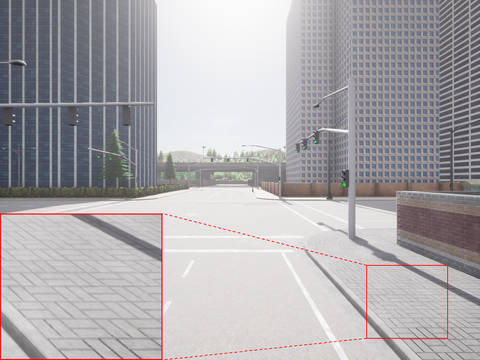}
    \captionsetup{labelformat=empty}
    \caption{HR}
\end{subfigure}
\hfill
\begin{subfigure}[b]{0.24\textwidth}
    \includegraphics[width=\textwidth]{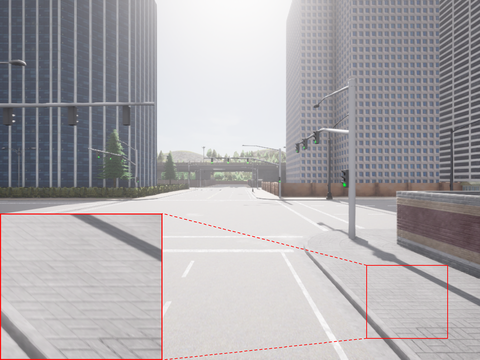}
    \captionsetup{labelformat=empty}
    \caption{Anime4K (FT)}
\end{subfigure}
\hfill
\begin{subfigure}[b]{0.24\textwidth}
    \includegraphics[width=\textwidth]{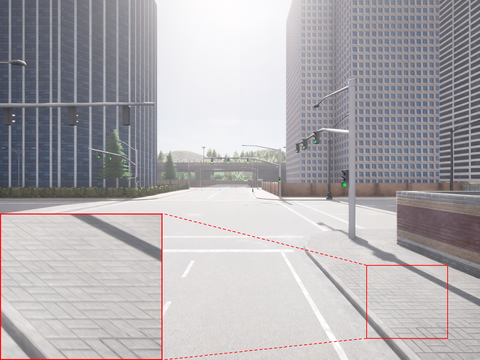}
    \captionsetup{labelformat=empty}
    \caption{AdaCode (FT)}
\end{subfigure}
\hfill
\begin{subfigure}[b]{0.24\textwidth}
    \includegraphics[width=\textwidth]{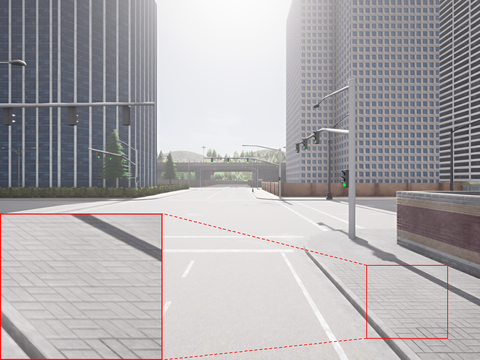}
    \captionsetup{labelformat=empty}
    \caption{Real-ESRGAN (FT)}
\end{subfigure}

\vspace{0.4cm} 

\begin{subfigure}[b]{0.24\textwidth}
    \includegraphics[width=\textwidth]{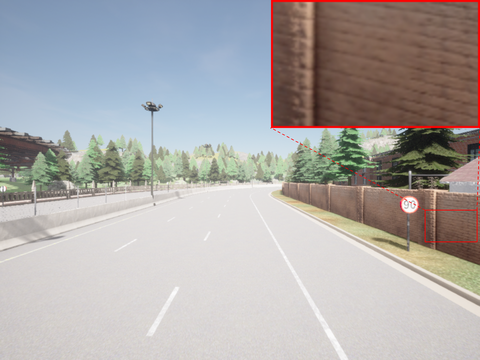}
    \captionsetup{labelformat=empty}
    \caption{LR}
\end{subfigure}
\hfill 
\begin{subfigure}[b]{0.24\textwidth}
    \includegraphics[width=\textwidth]{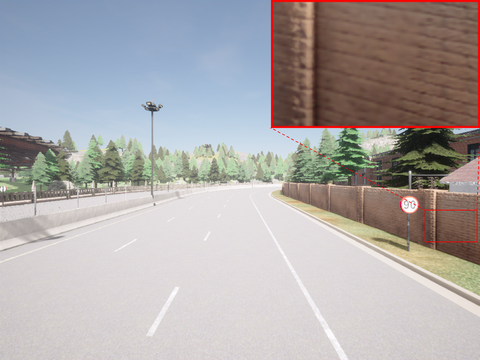}
    \captionsetup{labelformat=empty}
    \caption{\scriptsize Anime4K (PT)}
\end{subfigure}
\hfill
\begin{subfigure}[b]{0.24\textwidth}
    \includegraphics[width=\textwidth]{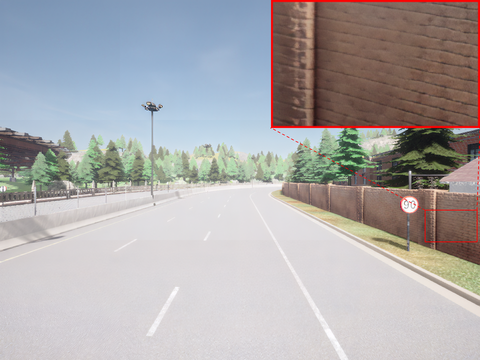}
    \captionsetup{labelformat=empty}
    \caption{AdaCode (PT)}
\end{subfigure}
\hfill
\begin{subfigure}[b]{0.24\textwidth}
    \includegraphics[width=\textwidth]{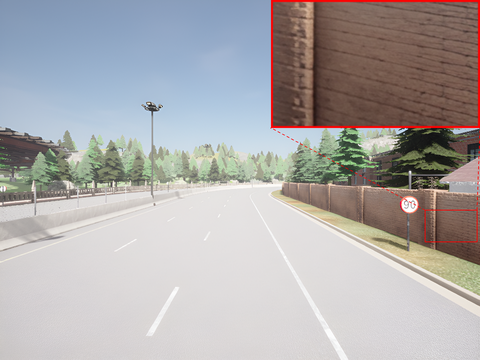}
    \captionsetup{labelformat=empty}
    \caption{Real-ESRGAN (PT)}
\end{subfigure}

\begin{subfigure}[b]{0.24\textwidth}
    \includegraphics[width=\textwidth]{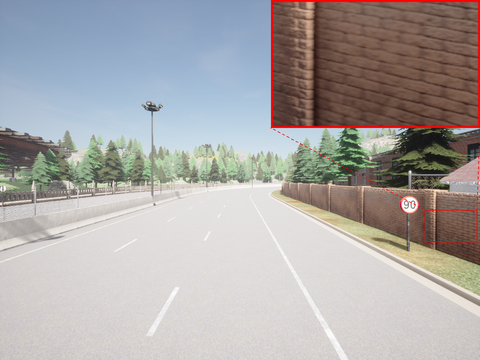}
    \captionsetup{labelformat=empty}
    \caption{HR}
\end{subfigure}
\hfill
\begin{subfigure}[b]{0.24\textwidth}
    \includegraphics[width=\textwidth]{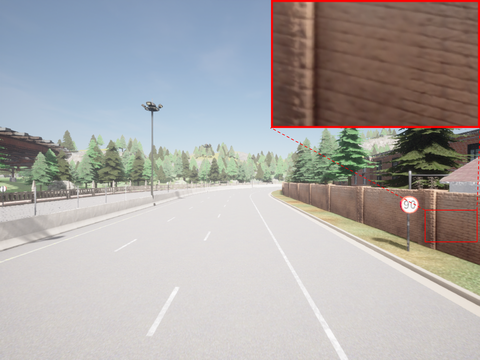}
    \captionsetup{labelformat=empty}
    \caption{Anime4K (FT)}
\end{subfigure}
\hfill
\begin{subfigure}[b]{0.24\textwidth}
    \includegraphics[width=\textwidth]{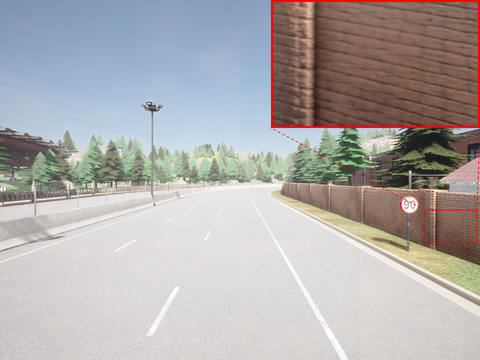}
    \captionsetup{labelformat=empty}
    \caption{AdaCode (FT)}
\end{subfigure}
\hfill
\begin{subfigure}[b]{0.24\textwidth}
    \includegraphics[width=\textwidth]{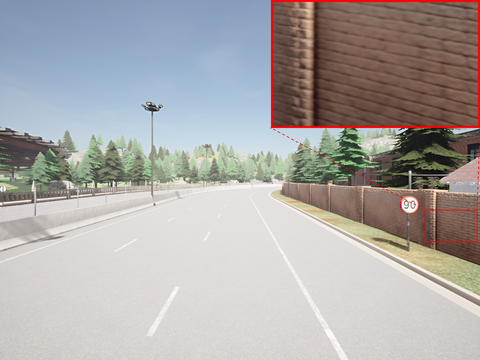}
    \captionsetup{labelformat=empty}
    \caption{Real-ESRGAN (FT)}
\end{subfigure}

\vspace{0.4cm} 

\begin{subfigure}[b]{0.24\textwidth}
    \includegraphics[width=\textwidth]{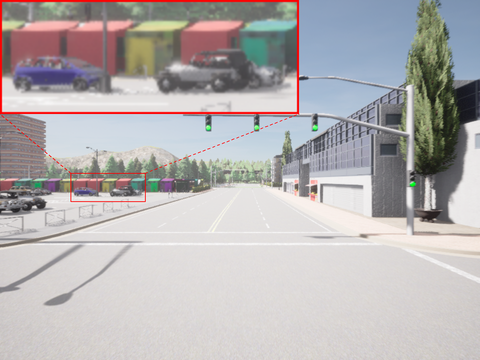}
    \captionsetup{labelformat=empty}
    \caption{LR}
\end{subfigure}
\hfill 
\begin{subfigure}[b]{0.24\textwidth}
    \includegraphics[width=\textwidth]{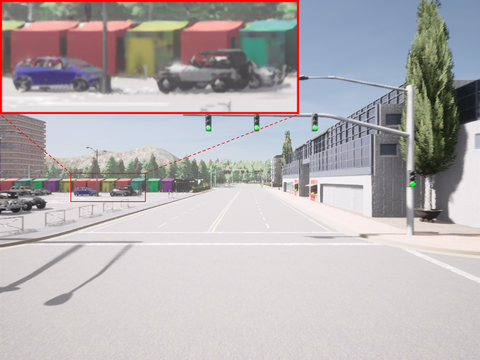}
    \captionsetup{labelformat=empty}
    \caption{\scriptsize Anime4K (PT)}
\end{subfigure}
\hfill
\begin{subfigure}[b]{0.24\textwidth}
    \includegraphics[width=\textwidth]{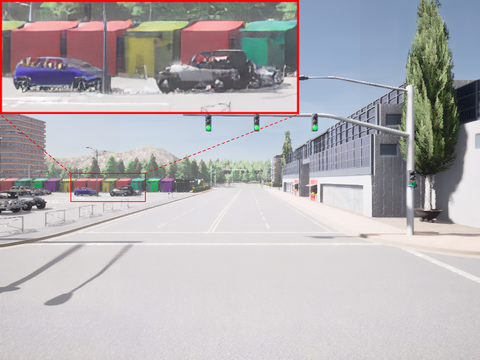}
    \captionsetup{labelformat=empty}
    \caption{AdaCode (PT)}
\end{subfigure}
\hfill
\begin{subfigure}[b]{0.24\textwidth}
    \includegraphics[width=\textwidth]{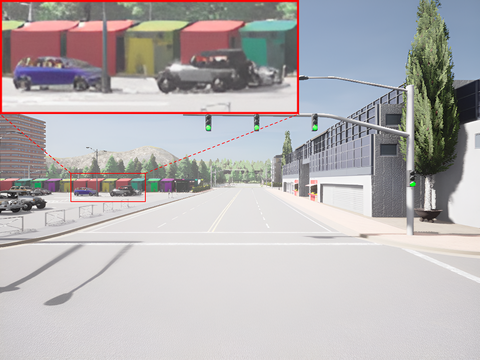}
    \captionsetup{labelformat=empty}
    \caption{Real-ESRGAN (PT)}
\end{subfigure}

\begin{subfigure}[b]{0.24\textwidth}
    \includegraphics[width=\textwidth]{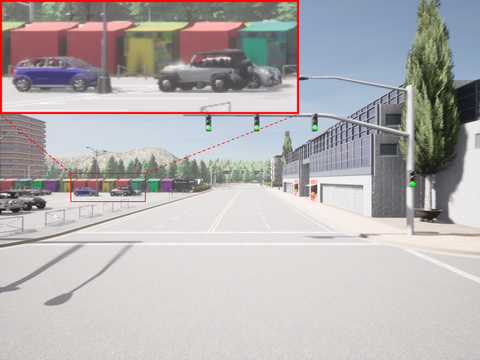}
    \captionsetup{labelformat=empty}
    \caption{HR}
\end{subfigure}
\hfill
\begin{subfigure}[b]{0.24\textwidth}
    \includegraphics[width=\textwidth]{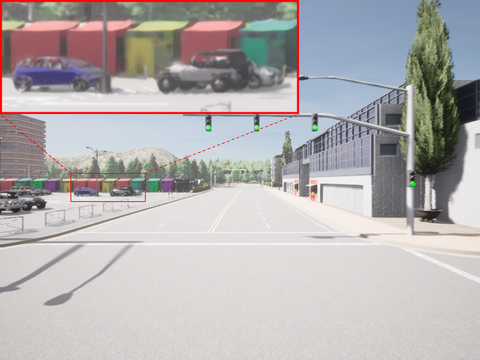}
    \captionsetup{labelformat=empty}
    \caption{Anime4K (FT)}
\end{subfigure}
\hfill
\begin{subfigure}[b]{0.24\textwidth}
    \includegraphics[width=\textwidth]{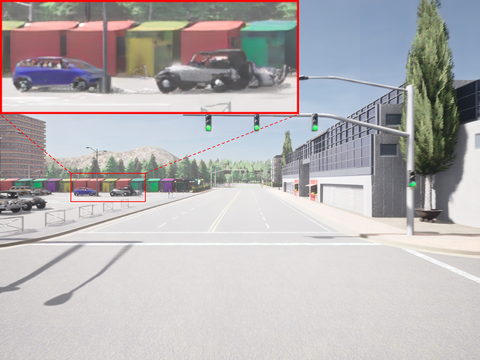}
    \captionsetup{labelformat=empty}
    \caption{AdaCode (FT)}
\end{subfigure}
\hfill
\begin{subfigure}[b]{0.24\textwidth}
    \includegraphics[width=\textwidth]{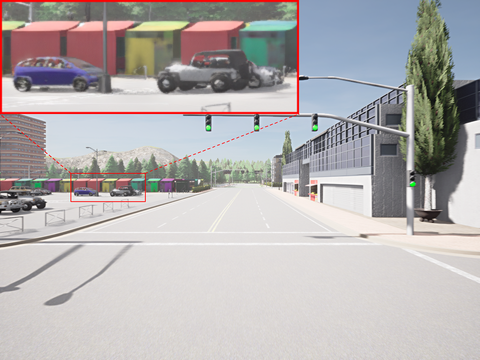}
    \captionsetup{labelformat=empty}
    \caption{Real-ESRGAN (FT)}
\end{subfigure}

\caption{Qualitative comparison of pretrained (PT) and finetuned (FT) super-resolution methods. Finetuning with ground-truth LR-HR paired data can restore intricate details more accurately with better clarity.}
\label{fig:sr_pretrain_and_finetuned_results}
\end{figure}

Without using GBuffer, we tested the pretrained models and finetuned models with the GameIR-SR training set.  Table~\ref{tab:sr_pretrain_and_finetuned_results} gives the quantitative performance comparison. Fig.~\ref{fig:sr_pretrain_and_finetuned_results} gives the qualitative comparison of example results. From the results, we can see that, after finetuning over the GameIR-SR training set, compared to the pretrained models, all three tested methods have large performance improvements across all metrics. As clearly shown in Fig.~\ref{fig:sr_pretrain_and_finetuned_results}, the finetuned models are capable of restoring intricate image details more accurately than the pretrained models. The finetuned models generate images with enhanced sharpness and clarity, especially in regions with rich textures. These results demonstrate the importance of training over ground-truth LR-HR paired data, where models can better learn real degradation features that match the inference stage. 

To further prove our point, we also used  BSRGAN\cite{BSRGAN} as the degradation model to generate pseudo LR training data and finetuned the tested pretrained models using such pseudo pairs. The results are also shown in Table~\ref{tab:sr_pretrain_and_finetuned_results}, which clearly show that such pseudo training data can not give any performance gain. That is, improvements can not be obtained by simply increasing pseudo training data. Ground-truth LR-HR gaming data are necessary for effective training.



\subsection{Performance with Additional GBuffer}
We modified the Real-ESRGAN\cite{Realesrgan} to make use of GBuffers as additional information to help SISR. There are two common ways to include additional input information into a model: as additional inputs or as generation conditions. \vspace{-1.5em}

\begin{table}[htbp]
\centering
\renewcommand\arraystretch{1.5}
\tabcolsep=0.38cm
\begin{tabular}{c  c  c  c  c}
\toprule
Input Channels & PSNR$\uparrow$ & SSIM$\uparrow$ & FID$\downarrow$ & LPIPS$\downarrow$ \\
\midrule
RGB (pretrained models)          & 30.1798 & 0.9040 & 10.0374 & 0.0535  \\ 
\midrule
w/ S(concat)   & 30.1065 & 0.9026 & \textbf{9.9637}  & \textbf{0.0518} \\
w/ D(concat)   & \textbf{30.2236} & 0.9038 & 10.2756 &\textbf{0.0499} \\
w/ SD(concat)  & 30.1639 & 0.9033 & \textbf{9.9164}  & \textbf{0.0524} \\
w/ S(SFT)      & 30.0750 & 0.9025 & 10.2182          & \textbf{0.0519}  \\
w/ D(SFT)      & 30.1049 & 0.9020  & \textbf{9.5882} & \textbf{0.0492}\\
w/ SD(SFT)     & \textbf{30.1996} & \textbf{0.9052} & \textbf{9.8511} & \textbf{0.0507} \\ 
\bottomrule
\end{tabular}
\caption{Performance of using different GBuffer information: segmentation map (S), depth map (D), or both (SD), by different methods: as additional input (concat), or as generative condition (SFT). 
}
\label{tab:sr_with_additional_gbffer_results}
\end{table}\vspace{-2.5em}

\noindent\textbf{GBuffer as Additional Input}. The GBuffers provide two types of information: segmentation maps and depth maps. We experimented with adding segmentation map or depth map alone, or adding both of them together. Specifically, we concatenated either the segmentation map or depth Map, or both of them with the RGB image to form a 4-channel or 5-channel input and adjusted the input layer of Real-ESRGAN accordingly. We followed the two-phase training procedure of the original Real-ESRGAN \cite{Realesrgan}. In phase 1, Real-ESRNet with RRDBNet was finetuned as the backbone using the GameIR-SR training data. In phase 2, we continued tuning with GAN loss, where Real-ESRNet served as the generator tuned together with a UNet discriminator. \vspace{.5em}

\noindent\textbf{GBuffer as Generation Condition}. The GBuffer information can serve as generation conditions during reconstruction. Following the two-phase training process of Real-ESRGAN, in phase 1, we added a control network (CtrlNet) \cite{ControlNet2023} to Real-ESRNet, feeding the segmentation map, depth map, or both of them into CtrlNet respectively. Then, we used a Spatial Feature Transform (SFT) module \cite{SFTGAN} to fuse the output of the Real-ESRNet and CtrlNet in reconstructing the final image output. CtrlNet comprises upsample layers and ResBlocks, which extract multi-scale features from segmentation maps or depth maps, or both. 
In the SFT module, the scale and shift layers use outputs from CtrlNet to perform spatial feature transformations on the feature maps. The scale layer dynamically adjusts the scales of the output feature maps by Real-ESRNet. The shift layer adjusts the baseline activation levels of the Real-ESRNet outputs, tuning feature activation by the conditional information. 

Table~\ref{tab:sr_with_additional_gbffer_results} gives the performance comparison of different ways of using GBuffer inputs. From the results, by using both the segmentation map and depth map as generative conditions, we can both improve distortion and perceptual quality across all metrics. Note that due to modifications of the models, no pretrained model is used here. Better performance can be expected in the future if larger-scale and more diverse gaming data can be available for learning a general pretrained model for general gaming content.

\section{Evaluation of NVS over GameIR-NVS}

We evaluated SOTA NeRF-based NVS algorithms over the GameIR-NVS dataset. For each town, we randomly selected 5 out of the 20 multiview videos with different spawn sites. For each selected video, we randomly sampled a subset of multiview frames to train NeRF-based models and then rendered the remaining frames to compare with the ground-truth frames for evaluation. 

We evaluated 4 latest methods: Instant-NGP\cite{Instant-NGP} that targets fast computation, NeRFacto\cite{nerfstudio} that improves efficiency in complex scenes, PyNeRF\cite{pynerf} that improves both speed and quality by training models across different spatial grid resolutions, and DSNerF\cite{DSNeRF} that uses depth maps to improve performance. 

We used the NeRFStudio implementation \cite{nerfstudio} of the tested methods. The default provided hyperparameters were used. The training and test were done on a single V100 graphics card.

We evaluated PSNR, SSIM, LPIPS, and FID to measure both distortion and visual quality. In addition, we evaluated normalized root mean square error (NRMSE) that measured the quality of generated depth maps. \vspace{-.5em}

\subsection{Performance without GBuffer} 

We tested 2 cases: using only the front view for both training and testing or using all 6 views (360°). In this experiment, for each video, we randomly extracted 10\% of the frames as the test set, with the remainder for training. Table~\ref{tab:nerf_results} gives the performance comparison.

The results show that PyNeRF outperforms the other methods across all metrics on our GameIR-NVS dataset. Also, metrics obtained using only front views are better than those using 360° views. We attribute this to the front view's single-directional nature as opposed to the comprehensive perspective provided by 360° views, simplifying the model's task by lowering the complexity involved in handling variations in lighting and perspective differences. 
\vspace{-.5em}


\begin{table}[htbp]
\renewcommand\arraystretch{1.5}
\begin{tabular}{ccccccccc}
\toprule
            & \multicolumn{2}{c}{PSNR$\uparrow$}            & \multicolumn{2}{c}{SSIM$\uparrow$}          & \multicolumn{2}{c}{FID$\downarrow$}           & \multicolumn{2}{c}{LPIPS$\downarrow$  }         \\
\midrule
            & front            & 360°             & front           & 360°            & front            & 360°             & front           & 360°            \\
Instant-NGP & 30.9481          & 27.7004          & 0.9373          & 0.9089          & 30.8528          & 44.7195          & 0.1095          & 0.1703          \\
NeRFacto    & 26.6457          & 26.7710          & 0.8539          & 0.8774          & 36.1718          & 43.0672          & 0.1206          & 0.1648          \\
PyNeRF      & \textbf{36.1329} & \textbf{32.0542} & \textbf{0.9612} & \textbf{0.9357} & \textbf{20.2392} & \textbf{36.3312} & \textbf{0.0758} & \textbf{0.1301} \\
\bottomrule
\end{tabular}
\caption{Performance of different NVS methods, using front views or 360° training data: PSNR and SSIM, the higher the better; LPIPS and FID, the lower the better.}
\label{tab:nerf_results}
\end{table}\vspace{-3.5em}

\subsection{Performance with Additional GBuffer}

\begin{figure}[ht]
    \centering
    \begin{tabular}{
        >{\centering\arraybackslash}m{0.001\textwidth} 
        >{\centering\arraybackslash}m{0.32\textwidth} 
        >{\centering\arraybackslash}m{0.32\textwidth}
        >{\centering\arraybackslash}m{0.32\textwidth}
    }
    & \textbf{Ground Truth} & \textbf{NeRFacto} & \textbf{Depth-NeRFacto} \\
    \rotatebox{90}{} & 
    \includegraphics[width=0.156\textwidth]{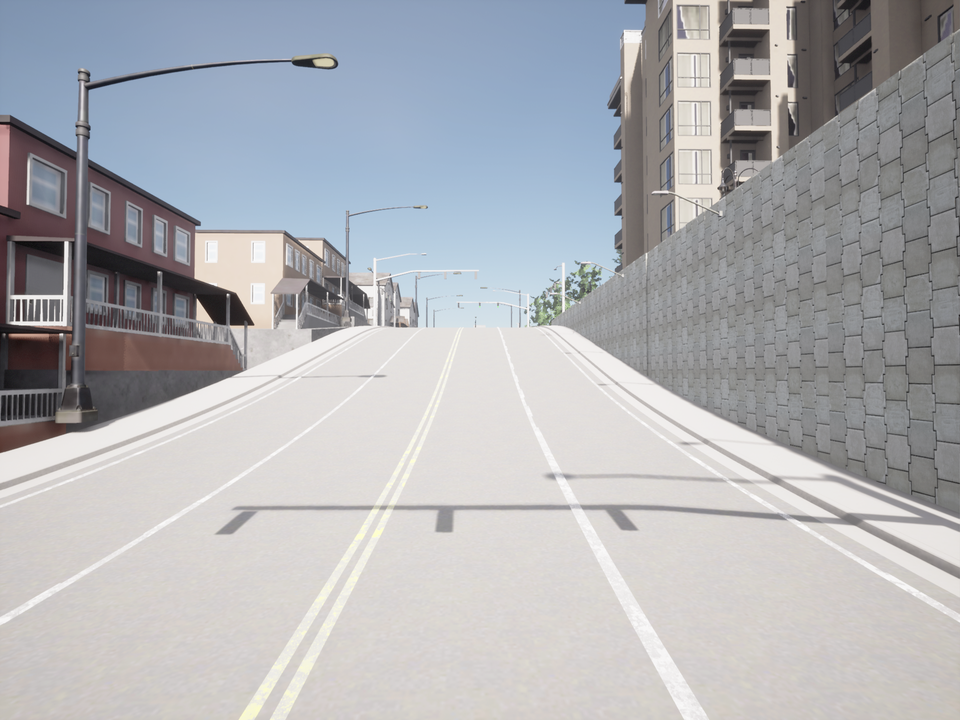} \hfill \includegraphics[width=0.156\textwidth]{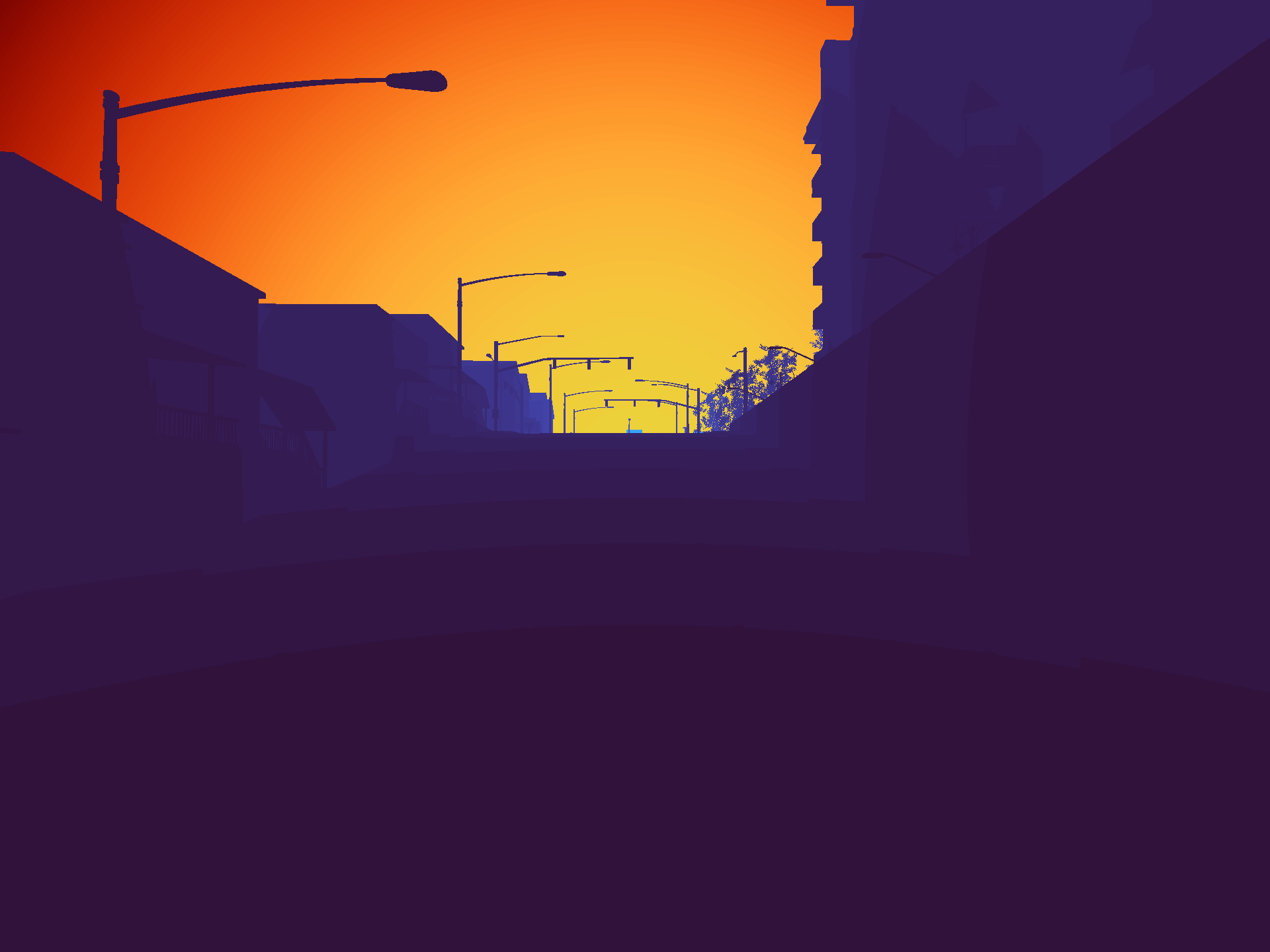} & 
    \includegraphics[width=0.156\textwidth]{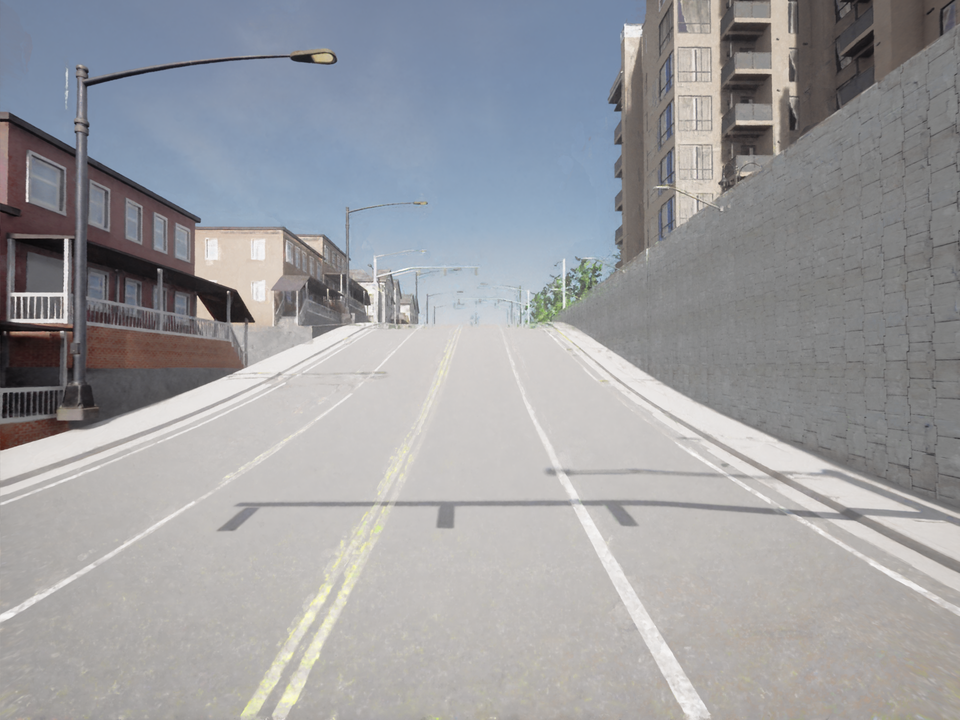} \hfill \includegraphics[width=0.156\textwidth]{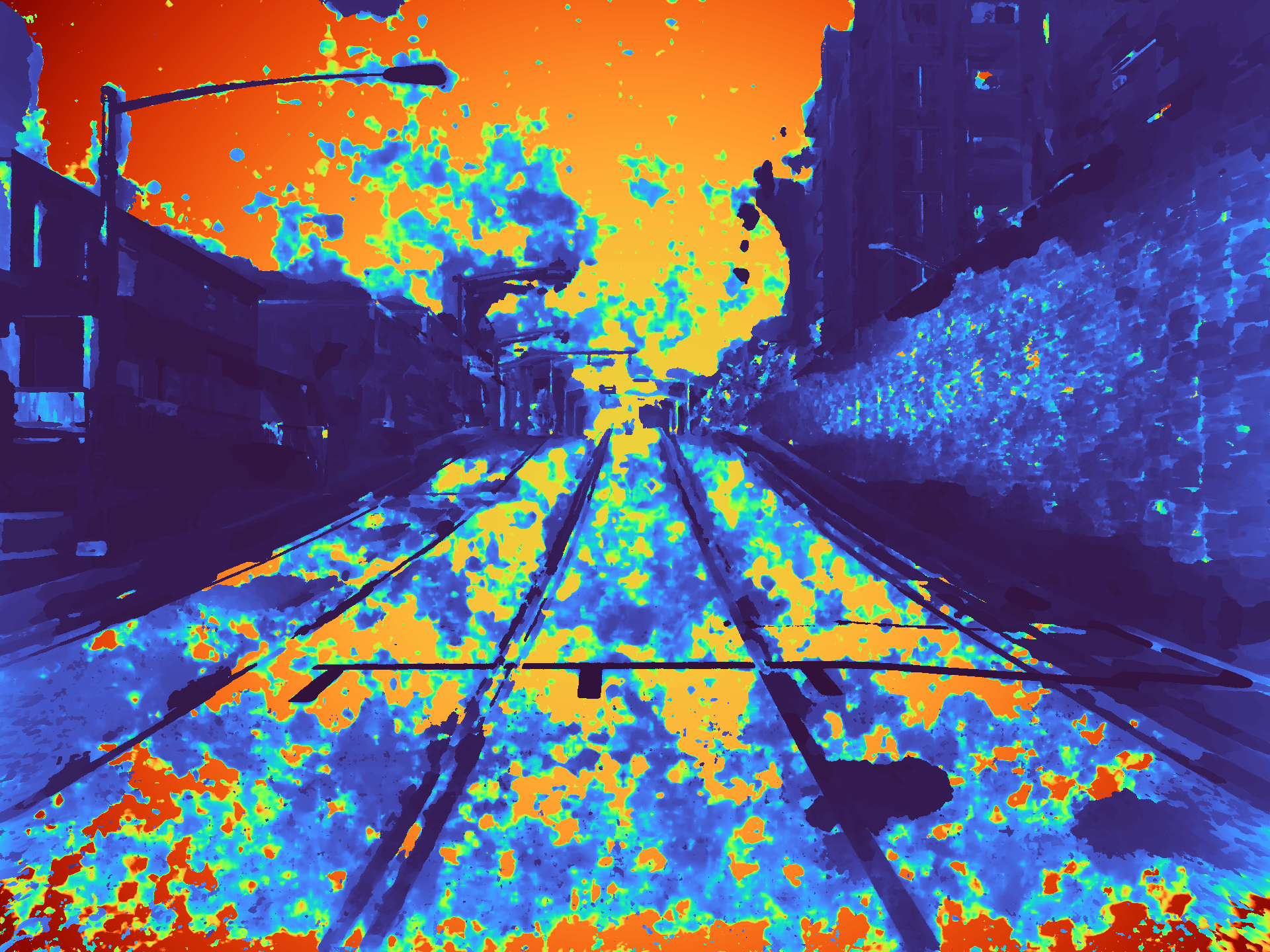} & 
    \includegraphics[width=0.156\textwidth]{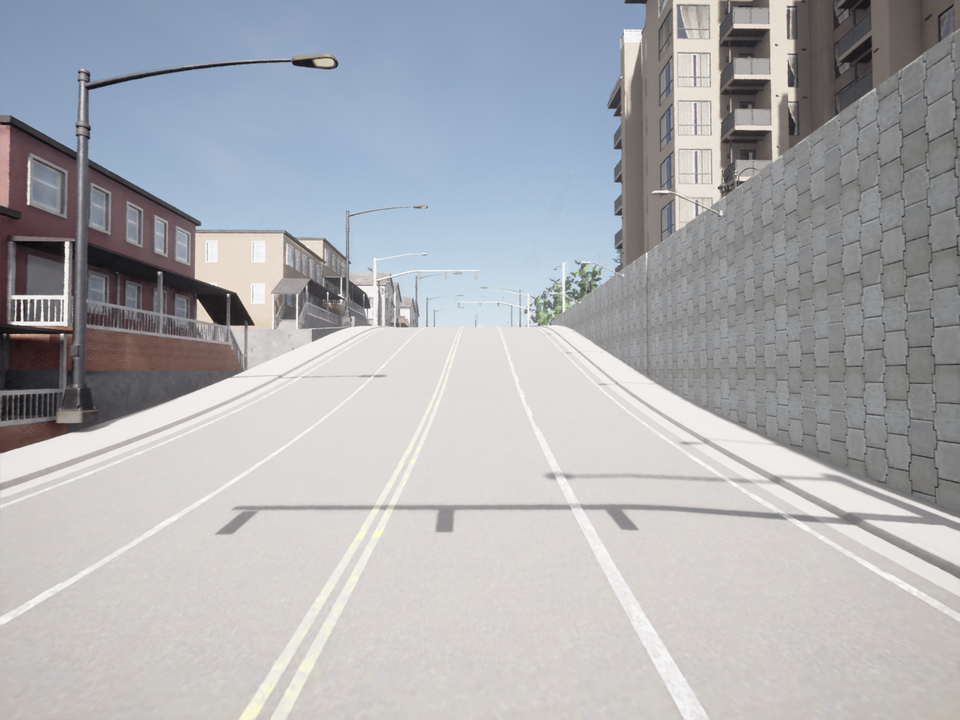} \hfill \includegraphics[width=0.156\textwidth]{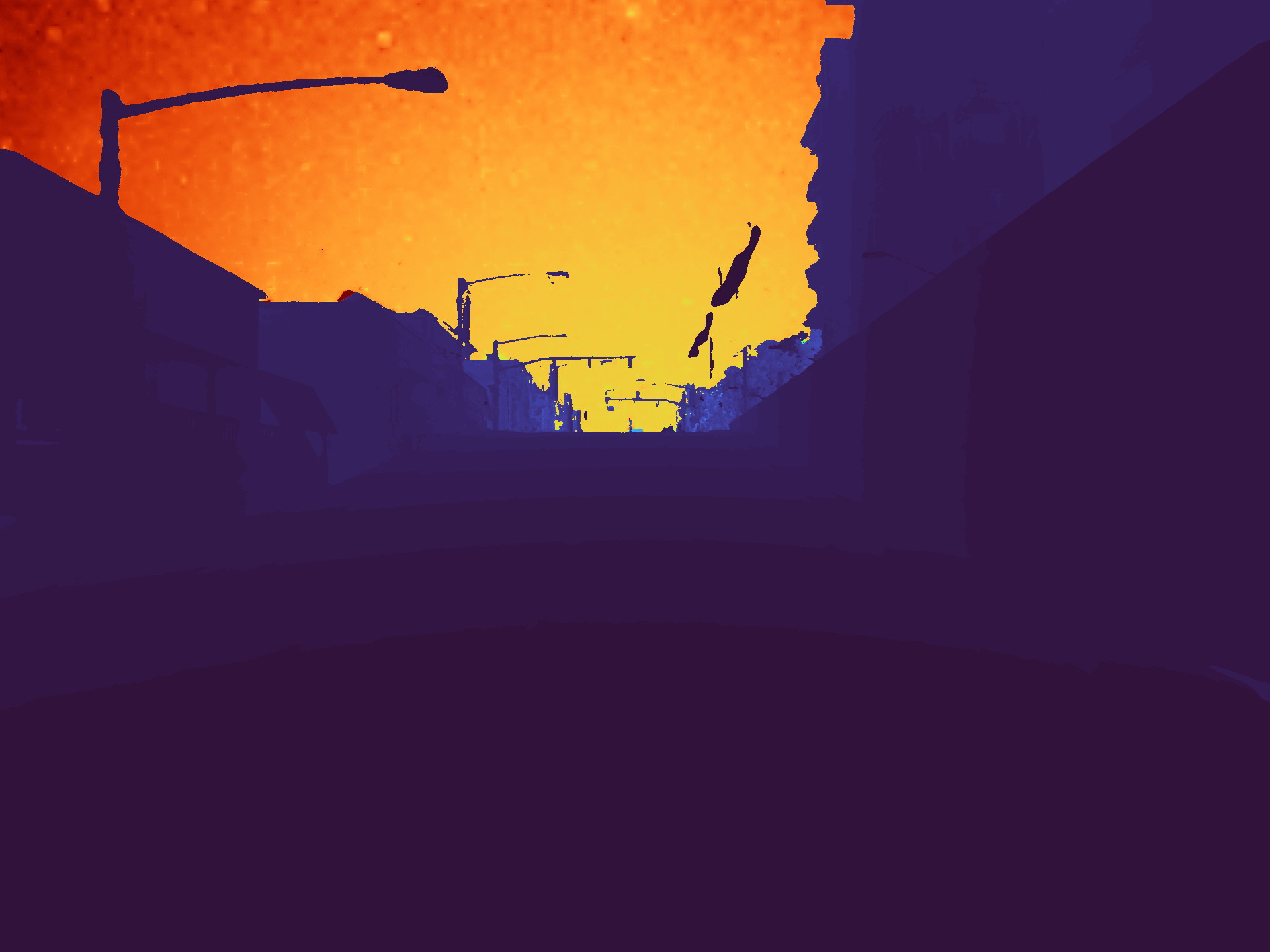} \\
    \rotatebox{90}{} & 
    \includegraphics[width=0.156\textwidth]{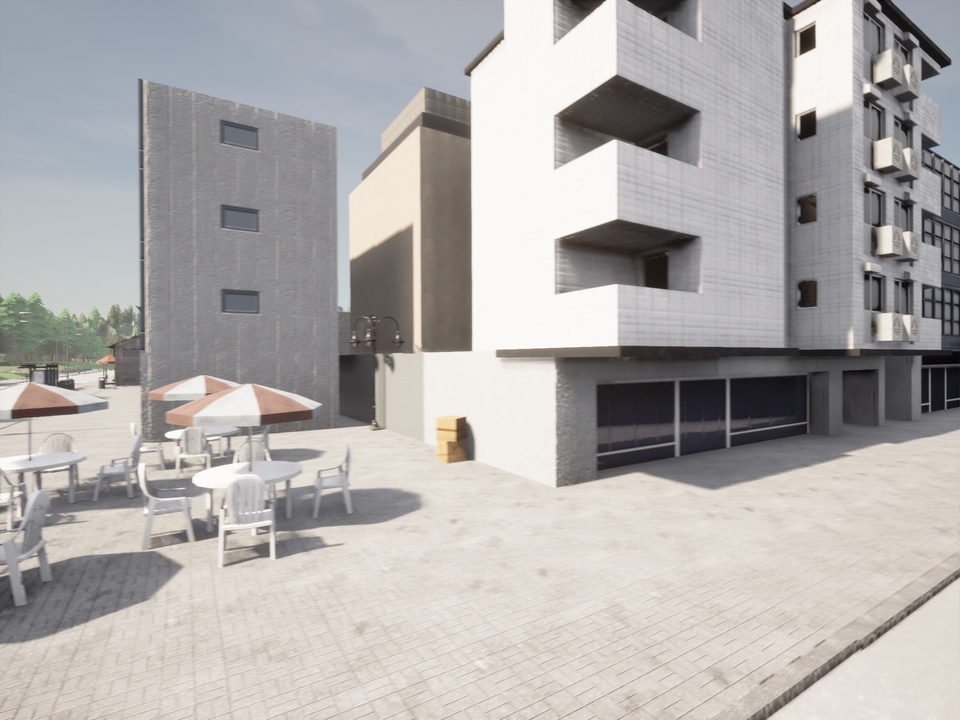} \hfill \includegraphics[width=0.156\textwidth]{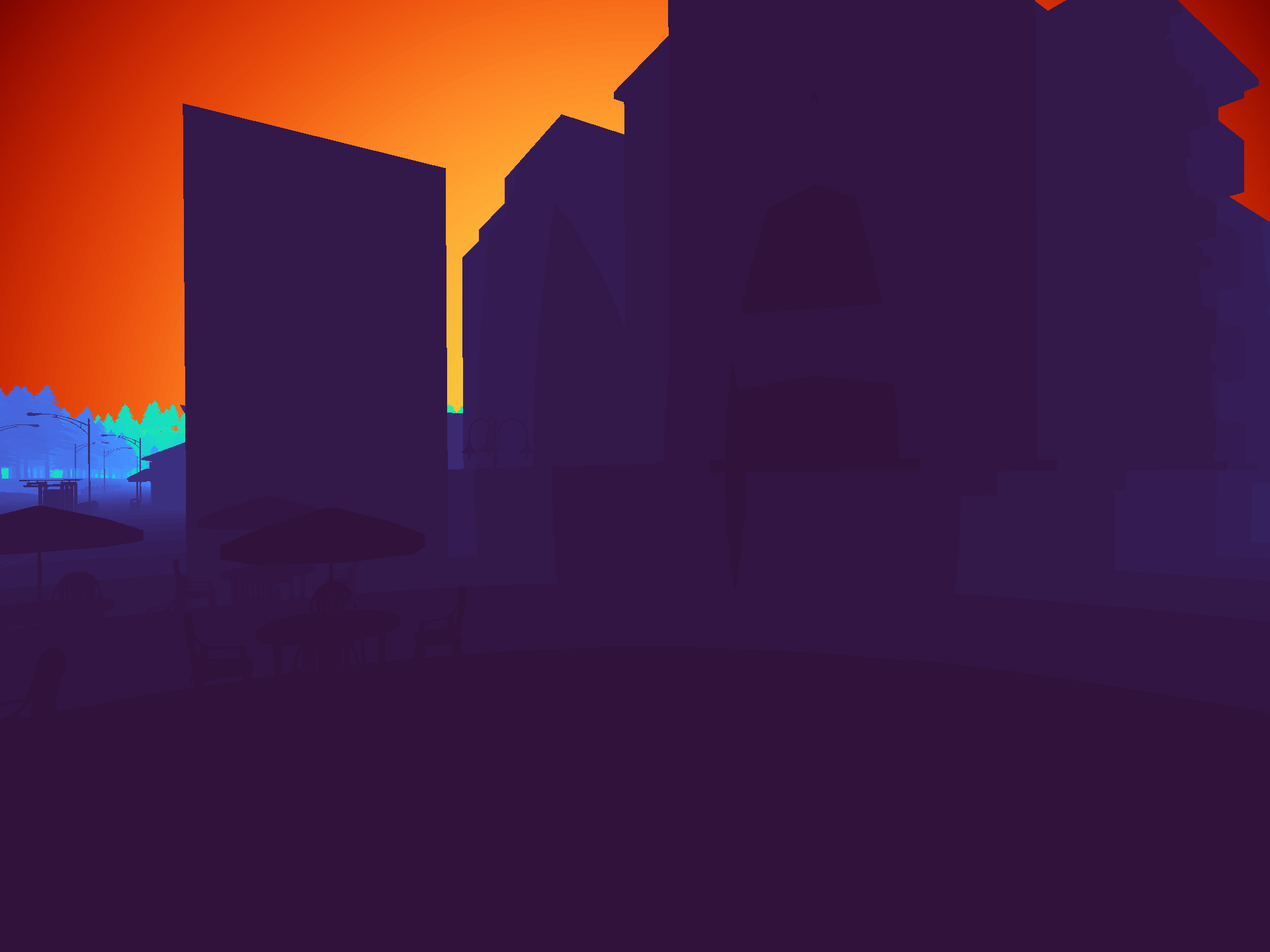} & 
    \includegraphics[width=0.156\textwidth]{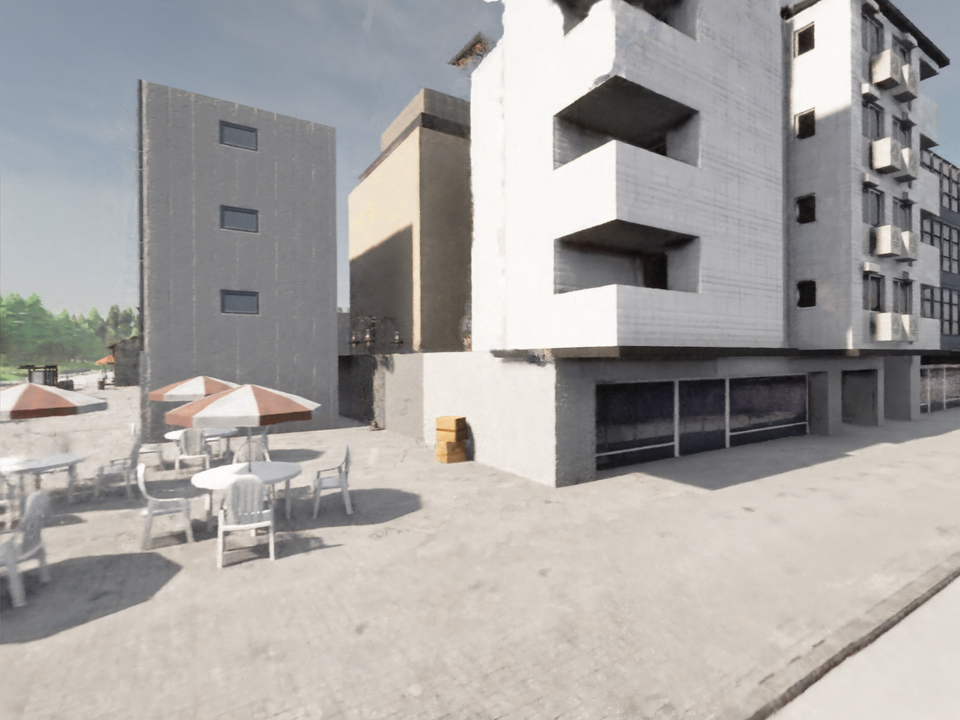} \hfill \includegraphics[width=0.156\textwidth]{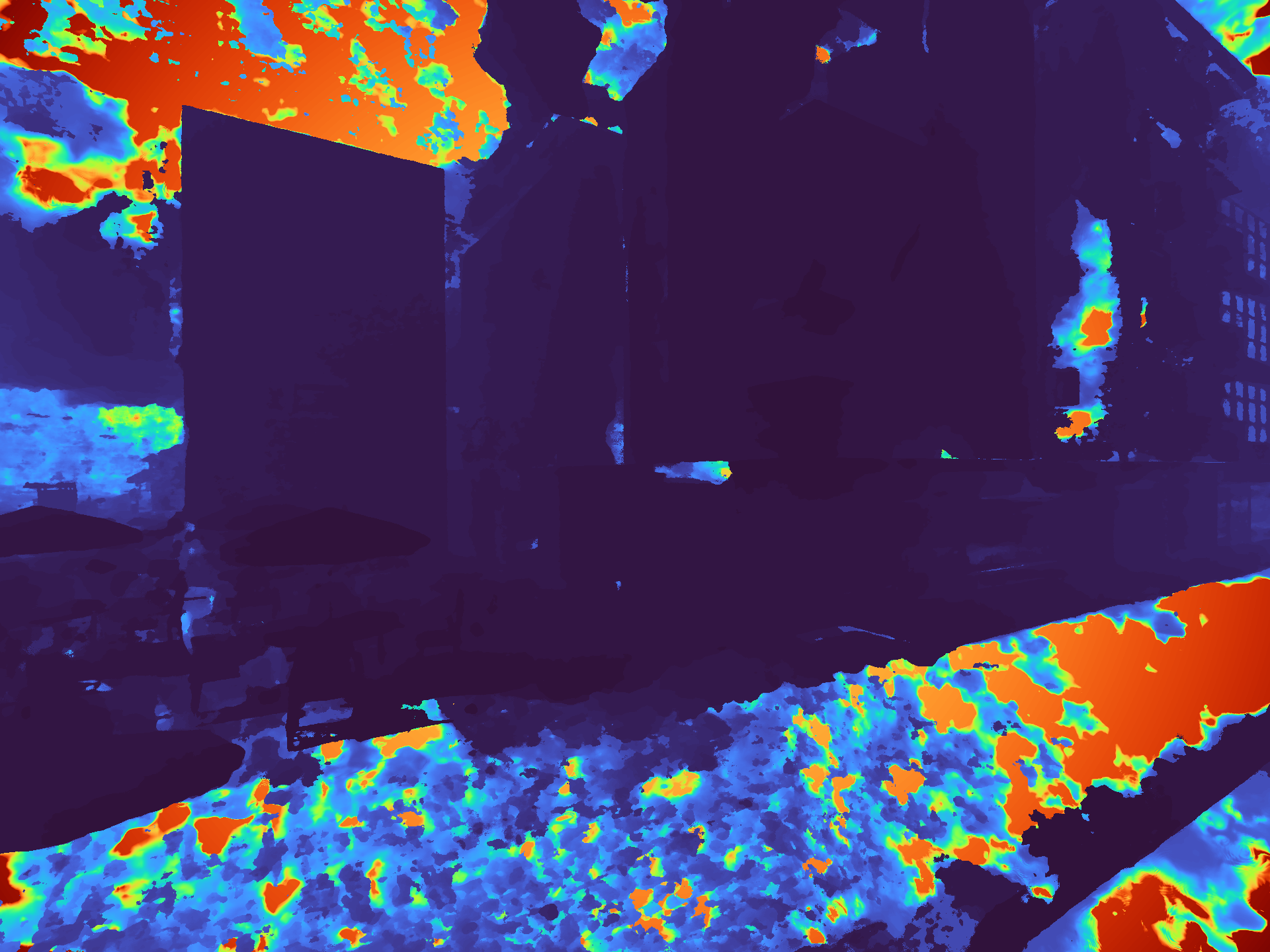} & 
    \includegraphics[width=0.156\textwidth]{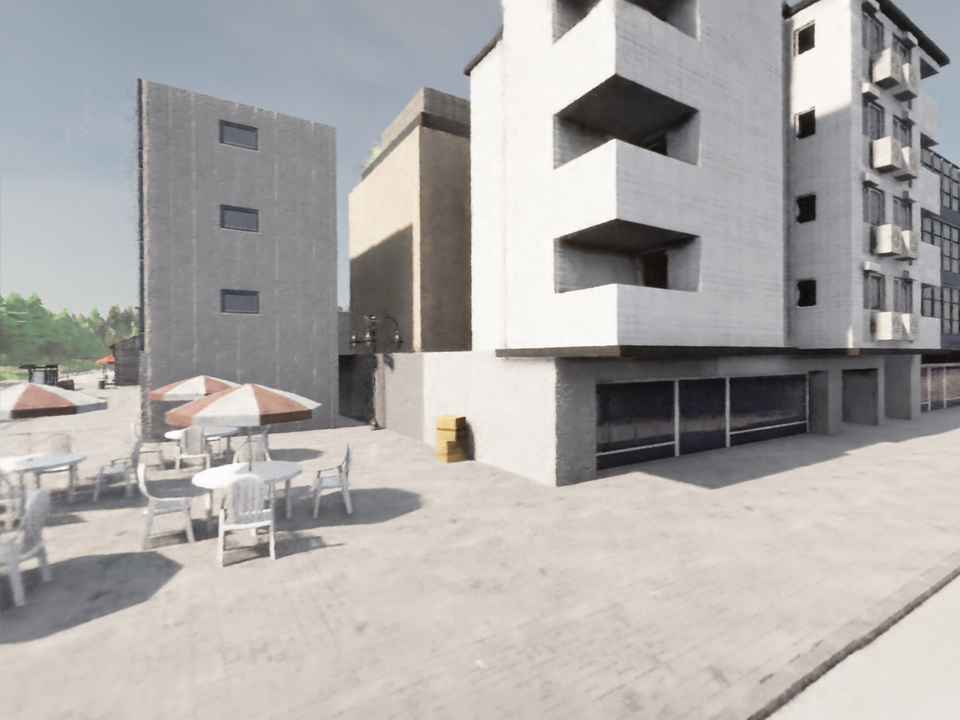} \hfill \includegraphics[width=0.156\textwidth]{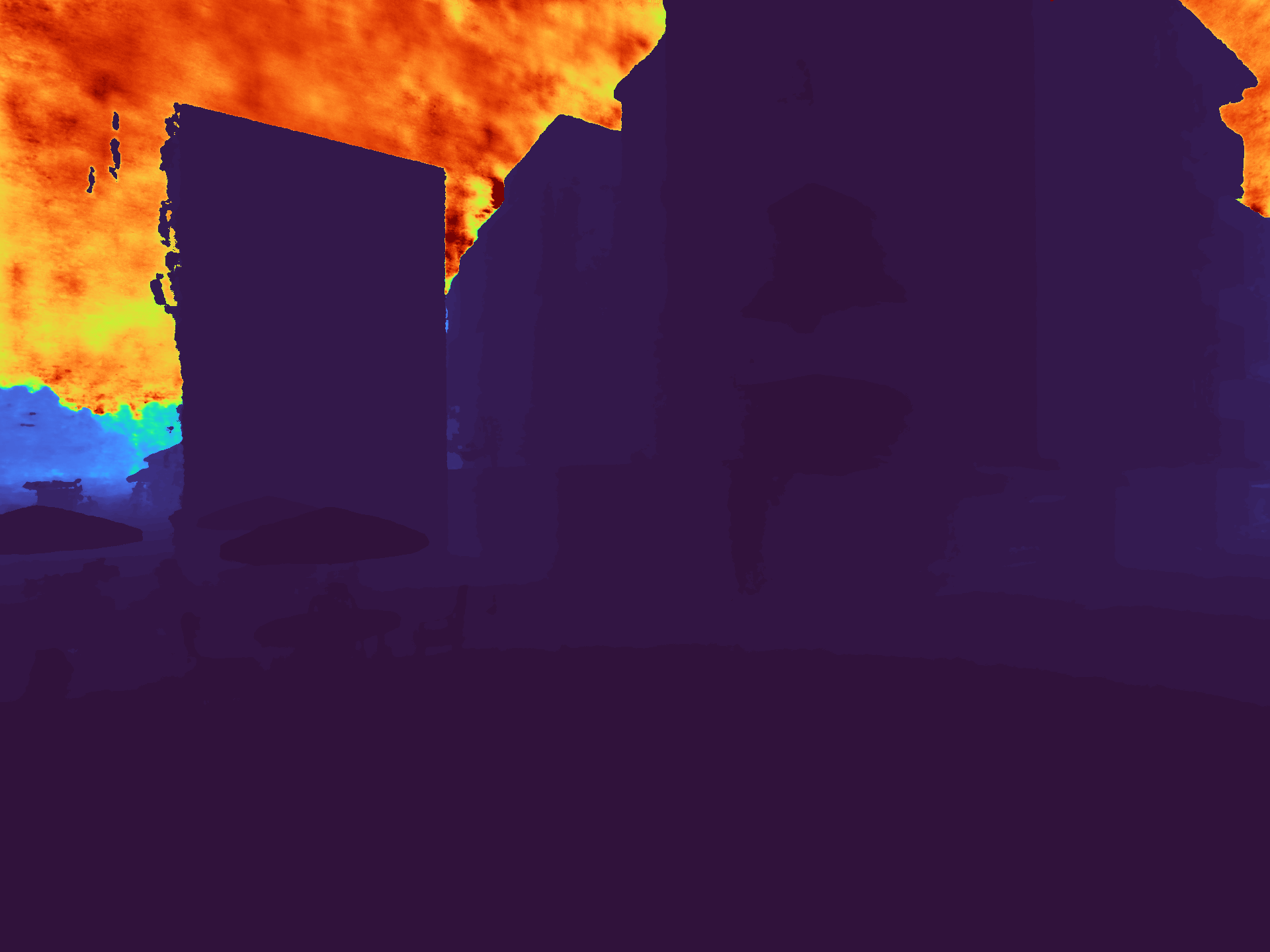} \\
    \rotatebox{90}{} & 
    \includegraphics[width=0.156\textwidth]{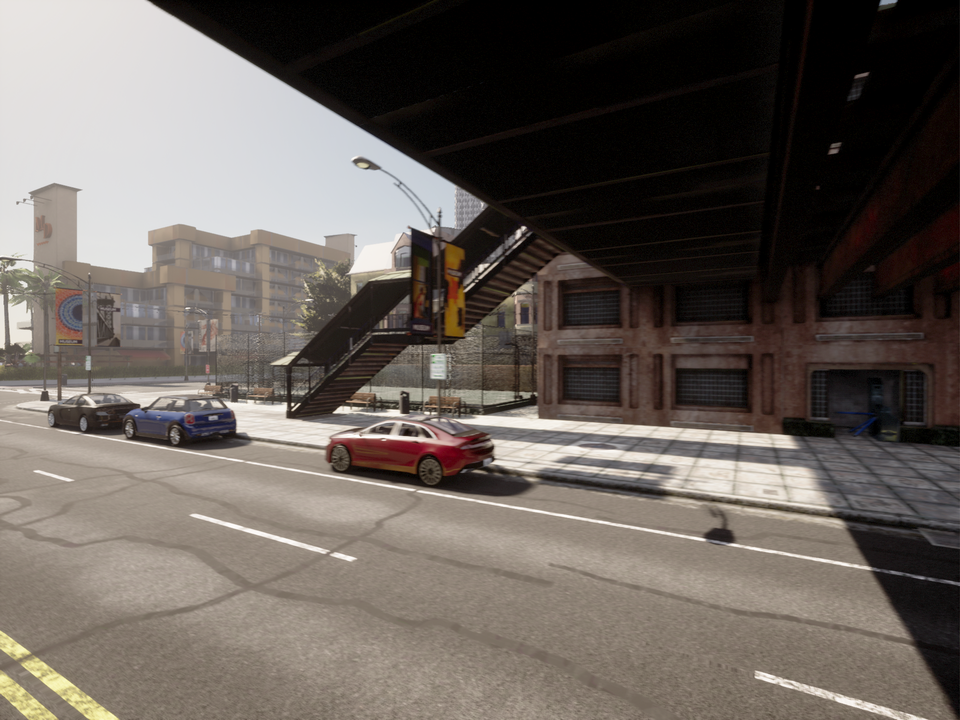} \hfill \includegraphics[width=0.156\textwidth]{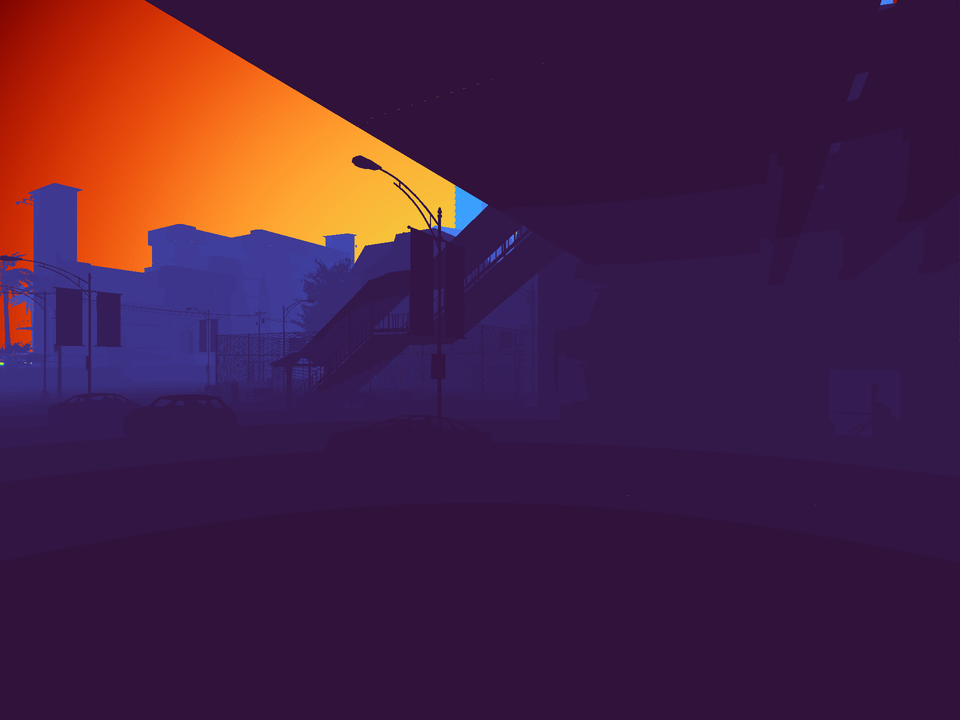} & 
    \includegraphics[width=0.156\textwidth]{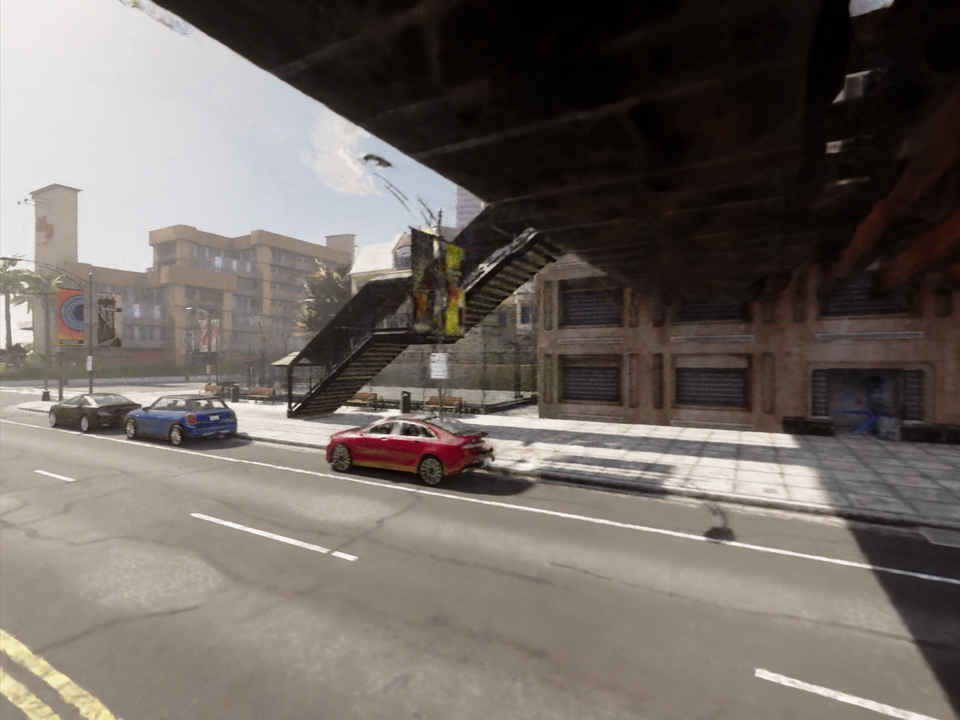} \hfill \includegraphics[width=0.156\textwidth]{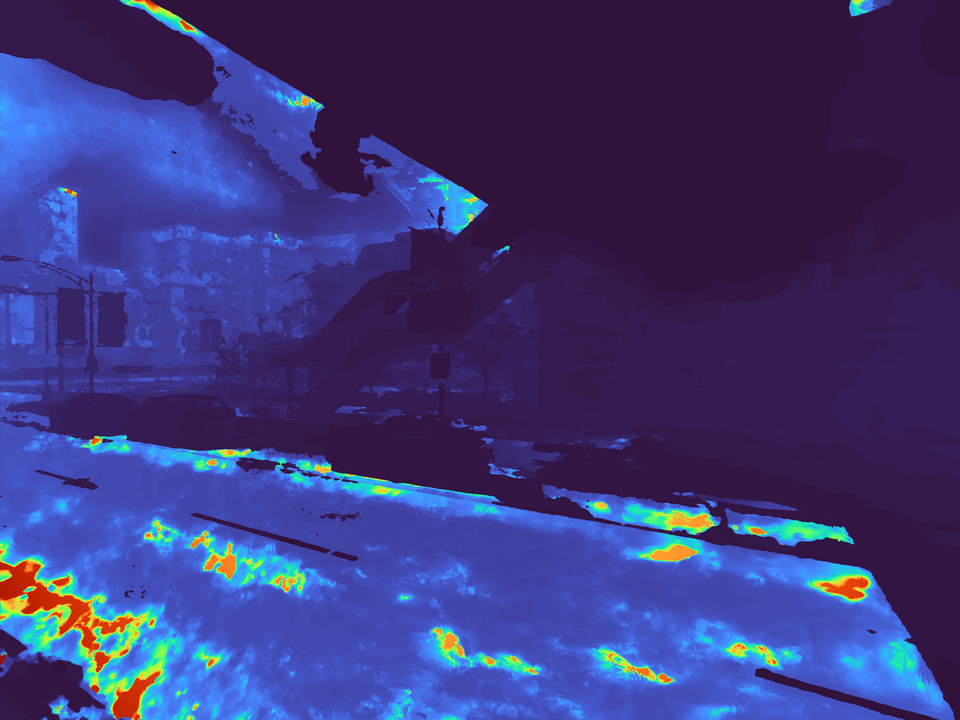} & 
    \includegraphics[width=0.156\textwidth]{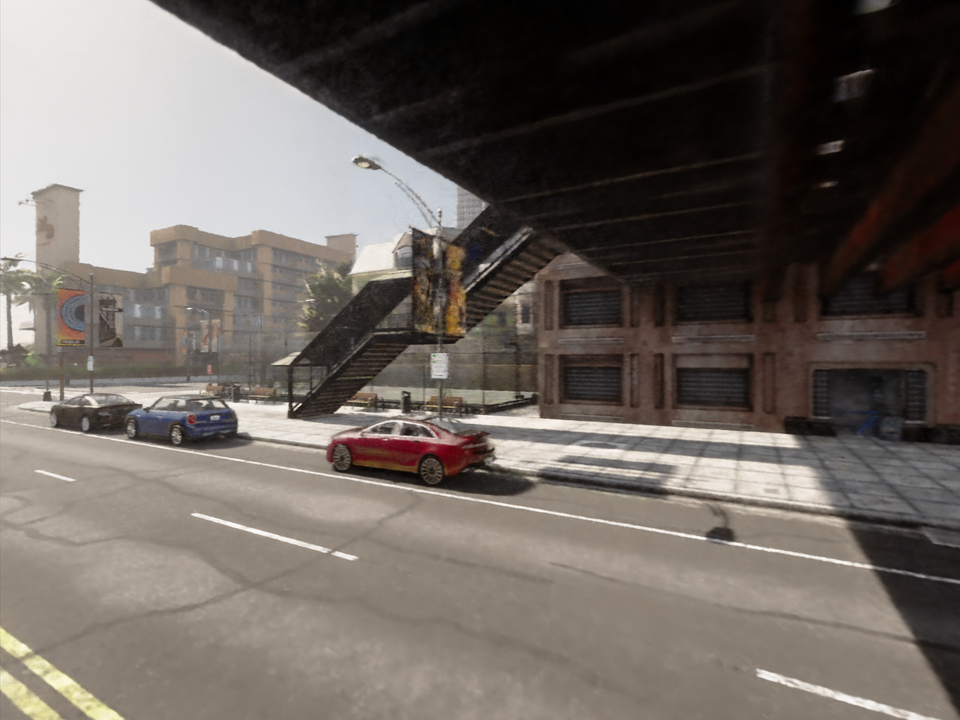} \hfill \includegraphics[width=0.156\textwidth]{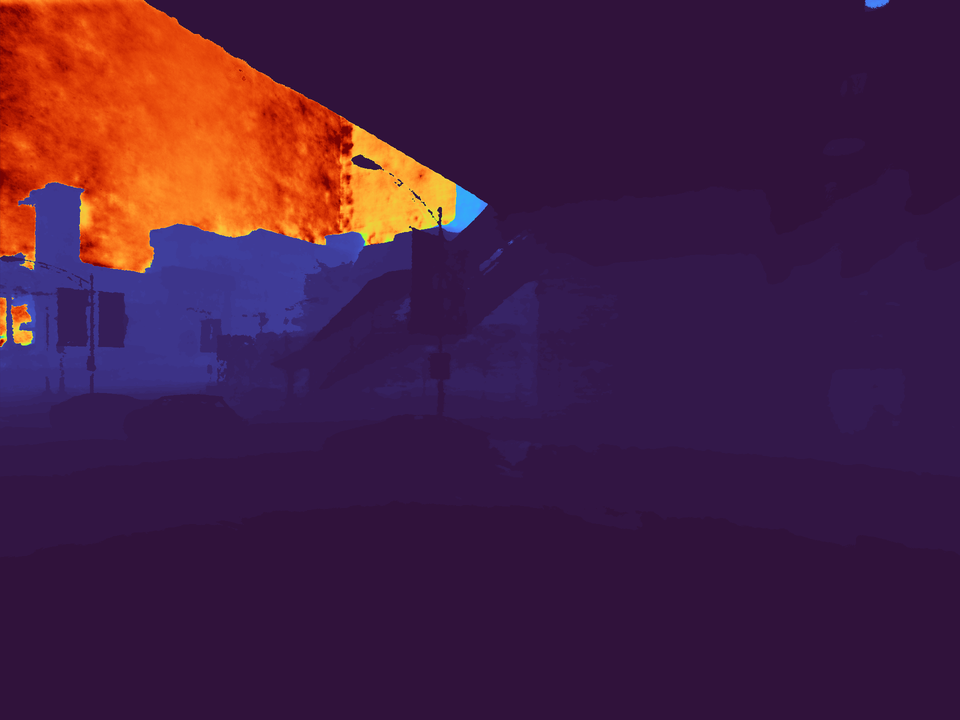} \\
    \end{tabular}
    \caption{Qualitative comparison of NeRFacto and Depth-NeRFacto. The RGB (left) and depth map (right) generated by Depth-NeRFacto better match the ground truth compared to NeRFacto.
    }
    \label{fig:ds_nerf_results}
\end{figure}

Motivated by the finding of DSNeRF \cite{DSNeRF} that supervising training with depth maps enhances the model's understanding of the scene, allowing it to render high-quality images with fewer training views, in this experiment, we aim to test how depth maps from GBuffer can help NVS with reduced training views. Specifically, we used NeRFactor and Depth-NeRFactor (their implementations in NeRFstudio \cite{nerfstudio}), where Depth-NeRFactor extended NeRFactor by incorporating depth maps as inputs and employed depth loss for supervised training. Instead of using 10\% frames for testing in the previous experiments, here we used 10\% randomly sampled frames for training, and the remaining 90\% for evaluation. Table~\ref{tab:ds_nerf_results} gives performance metrics, and Fig.~\ref{fig:ds_nerf_results} gives qualitative results.

\begin{table}[htbp]\scriptsize
\renewcommand\arraystretch{1.5}
\begin{tabular}{cccccccccll}
\toprule
               & \multicolumn{2}{c}{PSNR$\uparrow$}            & \multicolumn{2}{c}{SSIM$\uparrow$}          & \multicolumn{2}{c}{FID$\downarrow$}             & \multicolumn{2}{c}{LPIPS$\downarrow$}         & \multicolumn{2}{l}{Depth Err\%$\downarrow$}                      \\
\midrule
               & front            & 360°             & front           & 360°            & front            & 360°             & front           & 360°            & \multicolumn{1}{c}{front} & \multicolumn{1}{c}{360°} \\
NeRFacto       & 22.9597          & 23.3168          & 0.8133          & 0.8276          & 60.3013          & 40.8520          & 0.1654          & 0.1830          & 32.8053                   & 30.5997                  \\
\begin{tabular}[c]{@{}c@{}}Depth-\\ NeRFacto\end{tabular} & \textbf{25.3482} & \textbf{24.4990} & \textbf{0.8425} & \textbf{0.8306} & \textbf{29.3219} & \textbf{28.7427} & \textbf{0.1249} & \textbf{0.1736} & \textbf{5.4212}           & \textbf{7.2538}         \\
\bottomrule
\end{tabular}
\caption{Performance with 10\% training views: PSNR and SSIM, the higher the better; LPIPS, FID, and Depth Error, the lower the better.}
\label{tab:ds_nerf_results}
\end{table} 

As shown in Table~\ref{tab:ds_nerf_results}, using depth maps improves the NVS performance across all metrics. This demonstrates that depth maps can enrich the available geometric information even with limited training views, facilitating faster learning and a deeper understanding of the scene's 3D structure. Furthermore, Fig.~\ref{fig:ds_nerf_results} illustrates that depth maps enhance the realism of the generated RGB images, reduce distortions, and result in images with more detailed and textured representations. Results also clearly show that training with depth maps can significantly reduce depth error, and the depth maps generated by Depth-NeRFactor are very close to the ground truth. Such improvement is quite meaningful for gaming scenarios, as accurately generated depth maps provide precise three-dimensional geometric information. Such information not only enhances the visual authenticity of the game, but also improves collision detection, increasing the accuracy of physical interactions. This enables developers to handle collisions with complex-shaped objects more effectively, enhancing the overall gaming experience.

\vspace{-2em}

\section{Conclusion}
In this paper, we proposed GameIR, a large-scale synthetic dataset specifically designed for image restoration in gaming content. This dataset comprises two subsets aiming at two tasks: the GameIR-SR dataset for super-resolution and the GameIR-NVS dataset for NVS. GameIR-SR contains ground-truth LR and HR image pairs, and GameIR-NVS contains multiview videos with associated camera parameters. The corresponding GBuffers from deferred rendering are also provided for both datasets. We evaluated several SOTA algorithms for super-resolution and for NVS on our dataset to establish a baseline assessment for subsequent research on real gaming data. Additionally, we explored methods of utilizing GBuffers as supplementary information to help the super-resolution and NVS tasks. Our results demonstrated that GBuffers can provide enriched contextual information to improve performance.

This paper is our first attempt to provide ground-truth gaming data to facilitate public research on image restoration methods over the gaming content. In the future, we will continue to enrich our data collection by increasing the diversity of the data content, such as collecting different types of gaming data with different styles. 

\clearpage
%
%
\bibliographystyle{splncs04}
\bibliography{main}

\begin{thebibliography}{10}
\providecommand{\url}[1]{\texttt{#1}}
\providecommand{\urlprefix}{URL }
\providecommand{\doi}[1]{https://doi.org/#1}

\bibitem{DIV2K}
Agustsson, E., Timofte, R.: Ntire 2017 challenge on single image super-resolution: Dataset and study. In: The IEEE Conference on Computer Vision and Pattern Recognition (CVPR) Workshops (July 2017)

\bibitem{IRreview2}
Al-Mekhlafi, H., Liu, S.: Single image super-resolution: a comprehensive review and recent insight. Front. Comput. Sci. (181702) (2024)

\bibitem{Mip-NeRF}
Barron, J.T., Mildenhall, B., Tancik, M., Hedman, P., Martin-Brualla, R., Srinivasan, P.P.: Mip-nerf: A multiscale representation for anti-aliasing neural radiance fields. In: Proceedings of the IEEE/CVF International Conference on Computer Vision. pp. 5855--5864 (2021)

\bibitem{RealSR}
Cai, J., Zeng, H., Yong, H., Cao, Z., Zhang, L.: Toward real-world single image super-resolution: A new benchmark and a new model. In: Proceedings of the IEEE/CVF international conference on computer vision. pp. 3086--3095 (2019)

\bibitem{City100}
Chen, C., Xiong, Z., Tian, X., Zha, Z.J., Wu, F.: Camera lens super-resolution. In: Proceedings of the IEEE/CVF Conference on Computer Vision and Pattern Recognition. pp. 1652--1660 (2019)

\bibitem{Scannet}
Dai, A., Chang, A.X., Savva, M., Halber, M., Funkhouser, T., Nie{\ss}ner, M.: Scannet: Richly-annotated 3d reconstructions of indoor scenes. In: Proceedings of the IEEE conference on computer vision and pattern recognition. pp. 5828--5839 (2017)

\bibitem{Objaverse}
Deitke, M., Liu, R., Wallingford, M., Ngo, H., Michel, O., Kusupati, A., Fan, A., Laforte, C., Voleti, V., Gadre, S.Y., et~al.: Objaverse-xl: A universe of 10m+ 3d object. In: arXiv preprint arXiv:2307.05663 (2023)

\bibitem{ImageNet}
Deng, J., Dong, W., Socher, R., Li, L.J., Li, K., Fei-Fei, L.: Imagenet: A large-scale hierarchical image database. In: 2009 IEEE conference on computer vision and pattern recognition. pp. 248--255. Ieee (2009)

\bibitem{DSNeRF}
Deng1, K., Liu, A., Zhu, J., Ramanan, D.: Depth-supervised nerf: Fewer views and faster training for free. CVPR  (2022)

\bibitem{SRCNN1}
Dong, C., Loy, C.C., He, K., Tang, X.: Learning a deep convolutional network for image super-resolution. In: Computer Vision--ECCV 2014: 13th European Conference, Zurich, Switzerland, September 6-12, 2014, Proceedings, Part IV 13. pp. 184--199. Springer (2014)

\bibitem{SRCNN2}
Dong, C., Loy, C.C., He, K., Tang, X.: Image super-resolution using deep convolutional networks. IEEE transactions on pattern analysis and machine intelligence  \textbf{38}(2),  295--307 (2015)

\bibitem{CARLA}
Dosovitskiy, A., Ros, G., Codevilla, F., Lopez, A., Koltun, V.: {CARLA}: {An} open urban driving simulator. In: Proceedings of the 1st Annual Conference on Robot Learning. pp. 1--16 (2017)

\bibitem{elad1997restoration}
Elad, M., Feuer, A.: Restoration of a single superresolution image from several blurred, noisy, and undersampled measured images. IEEE transactions on image processing  \textbf{6}(12),  1646--1658 (1997)

\bibitem{Anime4K}
Feng, A.: Anime4k: A high-quality real time upscaler for anime video. \url{https://github.com/bloc97/Anime4K} (2019)

\bibitem{gu2019blind}
Gu, J., Lu, H., Zuo, W., Dong, C.: Blind super-resolution with iterative kernel correction. In: Proceedings of the IEEE/CVF conference on computer vision and pattern recognition. pp. 1604--1613 (2019)

\bibitem{DIV8K}
Gu, S., Lugmayr, A., Danelljan, M., Fritsche, M., Lamour, J., Timofte, R.: Div8k: Diverse 8k resolution image dataset. In: 2019 IEEE/CVF International Conference on Computer Vision Workshop (ICCVW). pp. 3512--3516 (2019). \doi{10.1109/ICCVW.2019.00435}

\bibitem{hevc_std}
{Int. Telecommun. Union-Telecommun. (ITU-T) and Int. Standards Org./Int/Electrotech. Commun. (ISO/IEC JTC 1)}: High efficiency video coding, rec. ITU-T H.265 and ISO/IEC 23008-2, 2019

\bibitem{vvc_std}
{ITU-T and ISO}: Versatile video coding, rec. ITU-T H.266 and ISO/IEC 23090-3, 2020

\bibitem{DTU}
Jensen, R., Dahl, A., Vogiatzis, G., Tola, E., Aanæs, H.: Large scale multi-view stereopsis evaluation. In: CVPR (2014)

\bibitem{johnson2016perceptual}
Johnson, J., Alahi, A., Fei-Fei, L.: Perceptual losses for real-time style transfer and super-resolution. In: Computer Vision--ECCV 2016: 14th European Conference, Amsterdam, The Netherlands, October 11-14, 2016, Proceedings, Part II 14. pp. 694--711. Springer (2016)

\bibitem{FFHQ}
Karras, T., Aila, T., Laine, S., Lehtinen, J.: Progressive growing of gans for improved quality, stability, and variation. arXiv preprint arXiv:1710.10196  (2017)

\bibitem{Splatting}
Kerbl, B., Kopanas, G., Thomas~Leimkuhler, Drettakis, G.: 3d gaussian splatting for real-time radiance field rendering. In: ACM Transactions on Graphics (ToG). vol.~42, pp. 1--14 (2023)

\bibitem{kim2016accurate}
Kim, J., Lee, J.K., Lee, K.M.: Accurate image super-resolution using very deep convolutional networks. In: Proceedings of the IEEE conference on computer vision and pattern recognition. pp. 1646--1654 (2016)

\bibitem{tanks-and-temples}
Knapitsch, A., Park, J., Zhou, Q.Y., Koltun, V.: Tanks and temples: Benchmarking large-scale scene reconstruction. ACM Transactions on Graphics (ToG)  \textbf{36}(4),  1--13 (2017)

\bibitem{IRreview3}
Lepcha, D., Goyal, B., Dogra, A., Goyal, V.: Image super-resolution: A comprehensive review, recent trends, challenges and applications. Information Fusion  \textbf{91},  230--260 (2023)

\bibitem{swinIR}
Liang, J., Cao, J., Sun, G., Zhang, K., Van-Gool, L., Timofte, R.: Swinir: Image restoration using swin transformer. arXiv preprint arXiv:2108.10257  (2021)

\bibitem{Flickr2K}
Lim, B., Son, S., Kim, H., Nah, S., Lee, K.M.: Enhanced deep residual networks for single image super-resolution. In: The IEEE Conference on Computer Vision and Pattern Recognition (CVPR) Workshops (July 2017)

\bibitem{lim2017enhanced}
Lim, B., Son, S., Kim, H., Nah, S., Mu~Lee, K.: Enhanced deep residual networks for single image super-resolution. In: Proceedings of the IEEE conference on computer vision and pattern recognition workshops. pp. 136--144 (2017)

\bibitem{COCO}
Lin, T.Y., Maire, M., Belongie, S., Hays, J., Perona, P., Ramanan, D., Doll{\'a}r, P., Zitnick, C.L.: Microsoft coco: Common objects in context. In: Computer Vision--ECCV 2014: 13th European Conference, Zurich, Switzerland, September 6-12, 2014, Proceedings, Part V 13. pp. 740--755. Springer (2014)

\bibitem{DL3DV-10K}
Ling, L., Sheng, Y., Tu, Z., Zhao, W., Xin, C., Wan, K., Yu, L., Guo, Q., Yu, Z., Lu, Y., Li, X., Sun, X., Ashok, R., Mukherjee, A., Kang, H., Kong, X., Hua, G., Zhang, T., Benes, B., Bera, A.: A large-scale scene dataset for deep learning-based 3d vision. In: CVPR (2024)

\bibitem{AdaCode}
Liu, K., Jiang, Y., Choi, I., Gu, J.: Learning image-adaptive codebooks for class-agnostic image restoration. ICCV  (2023)

\bibitem{DVC2019}
Lu, G., Ouyang, W., Xu, D., Zhang, X., Cai, C., Z.Gao: {DVC}: {A}n end-to-end deep video compression framework. CVPR pp. 11006--11015 (2019)

\bibitem{maeda2022image}
Maeda, S.: Image super-resolution with deep dictionary. In: European Conference on Computer Vision. pp. 464--480. Springer (2022)

\bibitem{michaeli2013nonparametric}
Michaeli, T., Irani, M.: Nonparametric blind super-resolution. In: Proceedings of the IEEE International Conference on Computer Vision. pp. 945--952 (2013)

\bibitem{NeRF}
Mildenhall, B., Srinivasan, P., Tancik, M., Barron, J., Ramamoorthi, R., Ng, R.: Nerf: Representing scenes as neural radiance fields for view synthesis. ECCV  (2020)

\bibitem{llff}
Mildenhall, B., Srinivasan, P.P., Ortiz-Cayon, R., Kalantari, N.K., Ramamoorthi, R., Ng, R., Kar, A.: Local light field fusion: Practical view synthesis with prescriptive sampling guidelines. ACM Transactions on Graphics (TOG)  (2019)

\bibitem{Instant-NGP}
M\"uller, T., Evans, A., Schied, C., Keller, A.: Instant neural graphics primitives with a multiresolution hash encoding. ACM Trans. Graph.  \textbf{41}(4),  102:1--102:15 (Jul 2022). \doi{10.1145/3528223.3530127}, \url{https://doi.org/10.1145/3528223.3530127}

\bibitem{NVDLSS1}
Nividia: Introducing nvidia dlss 3 \url{https://www.nvidia.com/en-us/geforce/news/dlss3-ai-powered-neural-graphics-innovations/}

\bibitem{NVDLSS3}
Nividia: Nvidia announces dlss 3.5 with ray reconstruction, boosting rt quality with an ai-trained denoiser. EuroGamer \url{https://www.eurogamer.net/digitalfoundry-2023-nvidia-announces-dlss-35-with-ray-reconstruction-boosting-rt-quality-with-an-ai-trained-denoiser}

\bibitem{NVDLSS2}
Nividia: Truly next-gen: Adding deep learning to games and graphics \url{https://www.gdcvault.com/play/1026184/Truly-Next-Gen-Adding-Deep}

\bibitem{D-nerf}
Pumarola, A., Corona, E., Pons-Moll, G., Moreno-Noguer, F.: D-nerf: Neural radiance fields for dynamic scenes. arXiv preprint arXiv:2011.13961  (2020)

\bibitem{VRreview1}
Rota, C., Buzzelli, M., Bianco, S., et~al.: Video restoration based on deep learning: a comprehensive survey. Artif Intell Rev  \textbf{56},  5317–5364 (2023)

\bibitem{Replica}
Straub, J., Whelan, T., Ma, L., Chen, Y., Wijmans, E., Green, S., Engel, J.J., Mur-Artal, R., Ren, C., Verma, S., et~al.: The replica dataset: A digital replica of indoor spaces. arXiv preprint arXiv:1906.05797  (2019)

\bibitem{nerfstudio}
Tancik, M., Weber, E., Ng, E., Li, R., Yi, B., Kerr, J., Wang, T., Kristoffersen, A., Austin, J., Salahi, K., Ahuja, A., McAllister, D., Kanazawa, A.: Nerfstudio: A modular framework for neural radiance field development. In: ACM SIGGRAPH 2023 Conference Proceedings. SIGGRAPH '23 (2023)

\bibitem{Mill-19}
Turki, H., Ramanan, D., Satyanarayanan, M.: Mega-nerf: Scalable construction of large-scale nerfs for virtual fly-throughs. In: Proceedings of the IEEE/CVF Conference on Computer Vision and Pattern Recognition (CVPR). pp. 12922--12931 (June 2022)

\bibitem{pynerf}
Turki, H., Zollh{\"o}fer, M., Richardt, C., Ramanan, D.: Pynerf: Pyramidal neural radiance fields. Advances in Neural Information Processing Systems  \textbf{36} (2024)

\bibitem{Realesrgan}
Wang, X., Xie, L., Dong, C., Shan, Y.: Real-esrgan: Training real-world blind super-resolution with pure synthetic data. In: International Conference on Computer Vision Workshops (ICCVW)

\bibitem{SFTGAN}
Wang, X., Yu, K., Dong, C., Loy, C.C.: Recovering realistic texture in image super-resolution by deep spatial feature transform. In: Proceedings of the IEEE conference on computer vision and pattern recognition. pp. 606--615 (2018)

\bibitem{wang2018esrgan}
Wang, X., Yu, K., Wu, S., Gu, J., Liu, Y., Dong, C., Qiao, Y., Change~Loy, C.: Esrgan: Enhanced super-resolution generative adversarial networks. In: Proceedings of the European conference on computer vision (ECCV) workshops. pp.~0--0 (2018)

\bibitem{ESRGAN}
Wang, X., Yu, K., Wu, S., Gu, J., Liu, Y., Dong, C., Qiao, Y., Change~Loy, C.: Esrgan: Enhanced super-resolution generative adversarial networks. In: Proceedings of the European conference on computer vision (ECCV) workshops. pp.~0--0 (2018)

\bibitem{DRealSR}
Wei, P., Xie, Z., Lu, H., Zhan, Z., Ye, Q., Zuo, W., Lin, L.: Component divide-and-conquer for real-world image super-resolution. In: Computer Vision--ECCV 2020: 16th European Conference, Glasgow, UK, August 23--28, 2020, Proceedings, Part VIII 16. pp. 101--117. Springer (2020)

\bibitem{Blendedmvs}
Yao, Y., Luo, Z., Li, S., Zhang, J., Ren, Y., Zhou, L., Fang, T., Quan, L.: Blendedmvs: A large-scale dataset for generalized multi-view stereo networks. In: Proceedings of the IEEE/CVF conference on computer vision and pattern recognition. pp. 1790--1799 (2020)

\bibitem{IRreview1}
Zhai, L., Wang, Y., Cui, S., Zhou, Y.: A comprehensive review of deep-learning-based real world image restoration. IEEE Access  (2023)

\bibitem{zhang2020deep}
Zhang, K., Gool, L.V., Timofte, R.: Deep unfolding network for image super-resolution. In: Proceedings of the IEEE/CVF conference on computer vision and pattern recognition. pp. 3217--3226 (2020)

\bibitem{BSRGAN}
Zhang, K., Liang, J., Van~Gool, L., Timofte, R.: Designing a practical degradation model for deep blind image super-resolution. In: IEEE International Conference on Computer Vision. pp. 4791--4800 (2021)

\bibitem{NeRF++}
Zhang, K., Riegler, G., Snavely, N., Koltun, V.: Nerf++: Analyzing and improving neural radiance fields. arXiv preprint arXiv:2010.07492  (2020)

\bibitem{zhang2017learning}
Zhang, K., Zuo, W., Gu, S., Zhang, L.: Learning deep cnn denoiser prior for image restoration. In: Proceedings of the IEEE conference on computer vision and pattern recognition. pp. 3929--3938 (2017)

\bibitem{zhang2018learning}
Zhang, K., Zuo, W., Zhang, L.: Learning a single convolutional super-resolution network for multiple degradations. In: Proceedings of the IEEE conference on computer vision and pattern recognition. pp. 3262--3271 (2018)

\bibitem{ControlNet2023}
Zhang, L., Rao, A., Agrawalau, M.: Adding conditional control to text-to-image diffusion models. ArXiv, arXiv:2302.05543  (2023)

\bibitem{zhang2018residual}
Zhang, Y., Tian, Y., Kong, Y., Zhong, B., Fu, Y.: Residual dense network for image super-resolution. In: Proceedings of the IEEE conference on computer vision and pattern recognition. pp. 2472--2481 (2018)

\bibitem{RealEstate10K}
Zhou, T., Tucker, R., Flynn, J., Fyffe, G., Snavely, N.: Stereo magnification: Learning view synthesis using multiplane images. arXiv preprint arXiv:1805.09817  (2018)

\end{thebibliography}
\end{document}